\documentclass[acmsmall]{acmart}

\AtBeginDocument{%
  \providecommand\BibTeX{{%
    \normalfont B\kern-0.5em{\scshape i\kern-0.25em b}\kern-0.8em\TeX}}}

\setcopyright{cc}
\setcctype{by}
\acmJournal{PACMHCI}
\acmYear{2025} \acmVolume{9} \acmNumber{7} \acmArticle{CSCW452} \acmMonth{11} \acmPrice{}\acmDOI{10.1145/3757633}

\usepackage[normalem]{ulem}
\usepackage{tcolorbox}
\usepackage{xcolor}
\usepackage{tabularx}

\begin{document}

\title[Investigating Social Influence of Multiple Agents in Human-Agent Interactions]{Multi-Agents are Social Groups: Investigating Social Influence of Multiple Agents in Human-Agent Interactions}


\author{Tianqi Song\textsuperscript{*}}
\affiliation{%
  \institution{National University of Singapore}
  \country{Singapore}
  }
\email{tianqi_song@u.nus.edu}
\orcid{0000-0001-6902-5503}

\author{Yugin Tan\textsuperscript{*}}
\affiliation{%
  \institution{National University of Singapore}
  \country{Singapore}
}
\email{tan.yugin@u.nus.edu}
\orcid{0009-0006-7357-0436}

\author{Zicheng Zhu}
\affiliation{%
  \institution{National University of Singapore}
  \country{Singapore}
}
\email{zicheng@u.nus.edu}
\orcid{0000-0002-4332-2515}

\author{Yibin Feng}
\affiliation{%
  \institution{National University of Singapore}
  \country{Singapore}
}
\email{feng.yibin@u.nus.edu}
\orcid{0009-0003-9109-8460}

\author{Yi-Chieh Lee}
\affiliation{%
  \institution{National University of Singapore}
  \country{Singapore}
}
\email{yclee@nus.edu.sg}
\orcid{0000-0002-5484-6066}

\renewcommand{\shortauthors}{Tianqi Song, Yugin Tan, Zicheng Zhu, Yibin Feng, and Yi-Chieh Lee}

\thanks{\textsuperscript{*}The two authors contributed equally to this work.}

\begin{teaserfigure}
  \centering
  \includegraphics[width=0.8\linewidth]{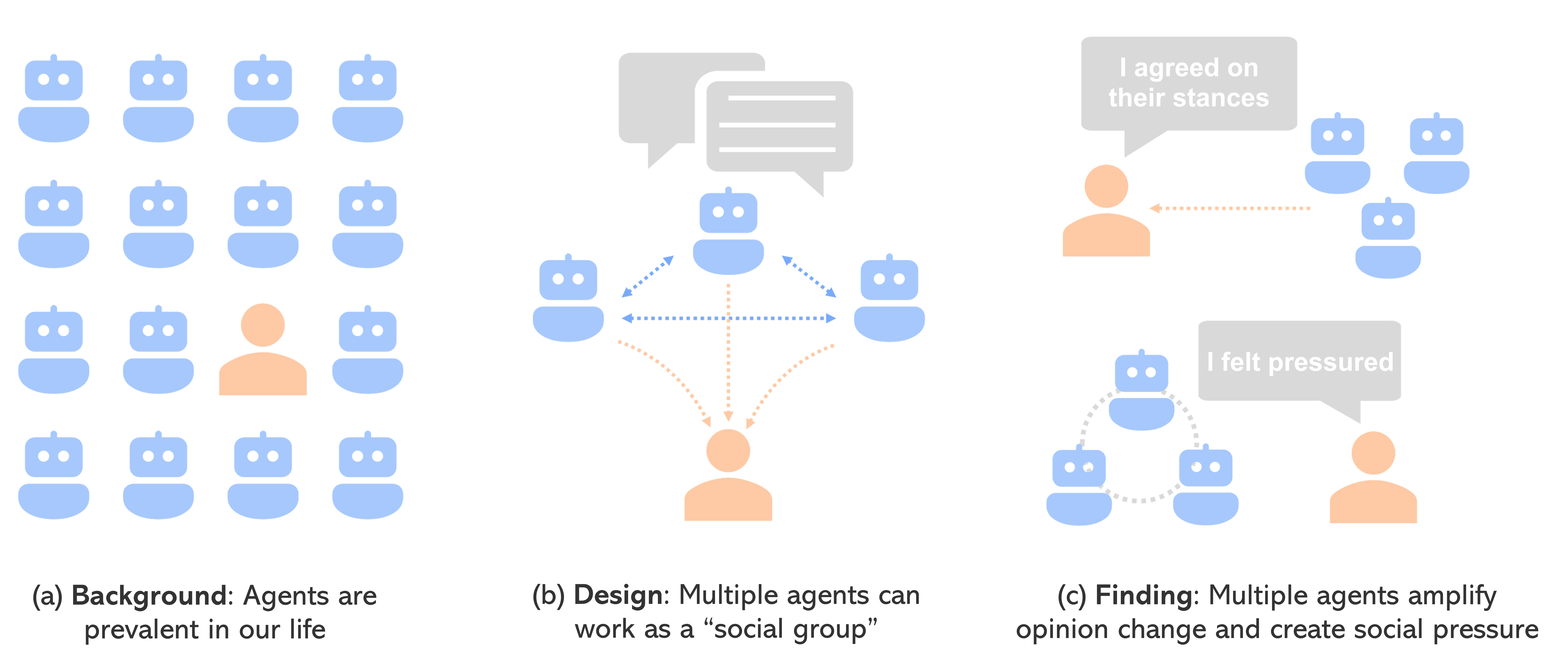}
  \caption{Overview of our study. (a) Agents are prevalent in our lives. They serve not only as tools but also can affect people's attitudes and behaviours. (b) When agents coordinate with each other and share the same stance, they can act as a "social group". (c) Participants changed more opinions and perceived stronger social pressure when interacting with multiple agents, compared with the single-agent setting.}
  \label{fig:teaser}
\end{teaserfigure}

\begin{abstract}
  Multi-agent systems, systems with multiple independent AI agents working together to achieve a common goal, are becoming increasingly prevalent in daily life.
Drawing inspiration from the phenomenon of human group social influence, we investigate whether a group of AI agents can create social pressure on users to agree with them, potentially changing their stance on a topic.
We conducted a study in which participants discussed social issues with either a single or multiple AI agents, and where the agents either agreed or disagreed with the user's stance on the topic. We found that conversing with multiple agents increased the social pressure felt by participants, and caused a greater shift in opinion towards the agents' stances on the conversation topics. Our study shows the potential advantages of multi-agent systems over single-agent platforms in causing opinion change. We discuss the resulting possibilities for multi-agent systems that promote social good, as well as potential malicious actors using these systems to manipulate public opinion.

\end{abstract}



\received{October 2024}
\received[revised]{April 2025}
\received[accepted]{August 2025}


\maketitle

\section{Introduction}

AI agents are advanced systems capable of interacting with users, holding conversations, and expressing their own thoughts \cite{wang2024survey}. These agents leverage the conversational and reasoning capabilities of large language models (LLMs) like Gemini, GPT-4, and LLaMA, and have become commonplace in daily life \cite{reuters2024openai}. Significant research has also explored the use of such agents in a broad range of scenarios, including education \cite{lieb2024student}, search interfaces \cite{ma2024beyond}, and trading platforms \cite{yu2024finmem}. The prevalence and ubiquity of LLM-powered agents have given rise to the concept of \textit{multi-agent} systems-systems with two or more independent agents working together to fulfill a common goal and improve task performance \cite{guo2024large, chen2023agentverse}. Major AI companies such as Anthropic \cite{anthropic2024claude} and Google \cite{google2024groopy} have introduced multi-agent frameworks, making it easier for developers to create such interfaces.

With both the rise of multi-agent systems and the increasing prevalence of virtual agents in our daily lives, an interesting question emerges: \textit{Can a group of AI agents collectively exert social influence on users?} This question arises from a well-documented observation in human-human interaction: individuals are more likely to be influenced by the opinions of larger groups of people, a phenomenon known as social influence. Various forms of social influence, including conformity to the majority \cite{cialdini2004social}, peer pressure \cite{shepherd2011susceptible, hui2009groupthink}, desire for social acceptance \cite{richters1991attachment, oostveen1996social}, and adherence to social norms \cite{mcdonald2015social, goldstein2011using}, are pervasive in everyday life.

While prior work in HCI has argued that human-computer interaction often mirrors human-human dynamics—where users perceive and respond to computers as social entities \cite{nass1994computers}—AI agents differ from humans in two important ways: 
First, regarding knowledge, AI agents are often perceived as having broader but more generic knowledge, particularly in factual or domain-specific areas, while lacking personal, contextual, or experiential insight \cite{luger2016like}. Unlike human groups, where informational influence tends to increase with group size \cite{cialdini2004social}, it is unclear whether users perceive a group of AI agents as more informational than a single agent. 
Second, in terms of social characteristics, AI agents often lack key social attributes \cite{ozdemir2023human} such as empathy, credibility, and authority, which are typically foundational to social influence in human groups \cite{cialdini2004social}. Consequently, it remains uncertain whether multiple AI agents can evoke the same sense of normative pressure or emotional affiliation that human groups do.
Given these differences, it remains an open question whether multiple AI agents can exert social influence in ways comparable to human social groups. This leads to a central research question of our work: Can a group of AI agents holding the same opinion influence user decisions in a manner similar to human group influence?

The importance of this question is twofold. Firstly, multi-agent interactions have become increasingly common, most notably on social media platforms like X (formerly Twitter), where human-like social bots have been effectively used to disseminate information and increase user engagement \cite{wischnewski2024agree}. Multi-agent systems have also been designed to support online discussions \cite{zhang2024see}, public information elicitation \cite{jiang2023communitybots}, and health coaching \cite{beinema2021tailoring}, exposing users to opinions from multiple agents in a range of settings. 
Secondly, the potential social influence of multi-agent systems on users would raise concerns. 
Evidence from psychology studies suggests that interactions with groups sharing similar views can significantly impact user opinions \cite{myers1976group, isenberg1986group}, potentially leading to opinion polarization and manipulation. 
As digital agents increasingly function as social actors, the influence they exert on users is a critical yet underexplored issue.

Beyond potential negative effects, the social influence of multi-agent systems can also be harnessed in a positive way. The social influence of majority opinion has long been used to encourage changes in attitudes and behaviors. In the traditional human-human domain, studies have shown their effectiveness in contexts such as health interventions \cite{skalski2007role, zhang2015leveraging} and sustainability efforts \cite{athanasiadis2005social, vossen2009social}. In human-computer interaction, persuasive systems have been designed across numerous domains like health and wellness \cite{balloccu2021unaddressed, oyebode2021tailoring}, education \cite{ahtinen2020learning, tian2021let}, and social welfare \cite{shi2020effects}, aiming to improve user behavior or change their attitudes through persuasion and informational education. Yet, these implementations are limited to single-agent interactions. The potential of multi-agent systems that collectively effect changes in user behavior or opinion remains unexplored.

The potential of AI agents in enacting persuasive effects is further supported by research in the HCI domain, which has highlighted the significant social influence individual AI agents exert, shaping users' attitudes \cite{jakesch2023co} and behaviors \cite{shi2020effects, zhang2020artificial}. However, such research has focused on the design characteristics of individual agents, such as how their perceived identity as a chatbot or human impacts outcomes \cite{shi2020effects}. Little attention has been given to how the \textit{number} of agents affects the effectiveness of these systems. In contrast, HCI research on multi-agent system research has largely revolved around enhancing task performance \cite{guo2024large, chen2023agentverse}, either by helping users complete tasks more effectively or improving the technical performance of the system itself \cite{du2023improving, chen2023reconcile}. The indirect effects of social influence caused by these multi-agent systems has been largely overlooked. The impact of social influence is less relevant to systems where the primary goal is to improve either the clarity of the agent's role in the discussion or the technical performance of the agents themselves. In scenarios where the system needs to both convey information and persuade a user of a certain viewpoint, the effect of multiple vs. single agents is critically underexplored.

Therefore, given 1) the increasing prevalence of multi-agent systems and interactions with multiple social bots online, 2) the known social influence that AI agents can exert, and 3) the established impacts of group social influence among humans that may be mirrored in human-AI interactions, there is 4) significant potential for both positive and negative effects caused by the collective social influence created by a group of AI agents. In this study, we address this issue by asking: \textit{Can interacting with multiple virtual agents create social influence that shifts our opinions?} 
Specifically, we explore whether interactions with multiple agents can lead to stronger opinion changes, induce varied social influence, and identify demographic factors that increase susceptibility to AI influence, aiming to inform ethical guidelines for multi-agent systems.

To investigate these questions, we conducted an experiment (\textit{n} = 94) where participants discussed their stances on two social topics with one, three, or five agents. Each participant was randomly assigned to one condition, with discussion content kept consistent across groups, differing only in the number of agents presenting it. Agents presented arguments that aligned with the participant’s stance on one topic and opposed it on the other. Survey data and open-ended responses were collected before and after each conversation round.

Through quantitative and qualitative analysis, we found that the \textbf{multi-agent setup significantly influenced opinion shifts}. Participants showed stronger opinion changes toward the agents when they disagreed and greater polarization when agents agreed with them. However, the shift did not intensify as the number of agents increased from 3 to 5. Instead, the 5-agent group led to a sense of polarization, with more participants rejecting the agents' arguments and moving further from the agents' positions. Additionally, \textbf{multi-agent groups heightened perceptions of social influence}, particularly \textbf{normative pressure} to align with the majority. Open-ended responses indicated that participants expressed a desire for affiliation within the group, which contributed to social influence and opinion shifts in the 3- and 5-agent conditions. Demographic analysis further revealed that younger participants were more susceptible to multi-agent influence.

Our study makes the following contributions to future research in the HCI and CSCW communities: 

\begin{itemize}
    \item {We show that having multiple agents in a discussion is more likely to create social influence and drive opinion change. This addresses a gap in understanding how multi-agent systems influence people, revealing their potential to drive stronger shifts in attitudes compared to single-agent setups.}
    \item{ We enhance the understanding of the differences in social influence between single-agent and multi-agent systems. By observing participants' feelings of social exclusion and their desire to engage, we identified an implicit sense of affiliation with a group of agents, expanding existing social influence theories to include human-agent interactions.}
    \item{ We provide design suggestions for using multi-agent setups to more effectively convey persuasive strategies. 
    By leveraging the normative pressure exerted by multiple agents, future systems could foster emotional engagement in participants, promoting more lasting and internalized changes in attitudes and behaviors.
    }
\end{itemize}

\section{Related Work}
\label{sec:relatedwork}

\subsection{Multi-agent Interaction}
\label{subsec:multiagentinteraction}

AI agents are increasingly integrated in our daily life. These advanced AI systems are capable of interacting with users, holding conversations, and expressing their own thoughts \cite{wang2024survey}. Recent research and industry applications have leveraged the conversational and reasoning capabilities of LLMs to create problem-solving tools \cite{wang2024executable, roy2024exploring, ning2024cheatagent, yu2024finmem, guan2024intelligent}, educational assistants \cite{lieb2024student}, and novel search interfaces \cite{ma2024beyond, yang2025understanding}. In commercial settings, many companies deploy AI agents on social media to assist with branding and customer service \cite{xu2017new, hu2018touch, jiang2022ai}, while others embed agents in their applications for assisting writing \cite{dhillon2024shaping}, brainstorming \cite{rezwana2023designing}, or gaming \cite{sidji2024human}.

Despite this, there are many tasks that single LLM-based agents struggle to handle effectively. Individual LLMs can lack sufficient domain expertise to handle specialized tasks \cite{ge2023advances}, and may have insufficiently robust reasoning capabilities. Inspired by existing psychological theories \cite{minsky1991society}, studies have investigated techniques such as "debating" \cite{du2023improving} or "reconciling" \cite{chen2023reconcile} between multiple agents, to enhance reasoning by leveraging diverse perspectives. This has driven growing interest in "multi-agent collaboration," where agents work together to improve task performance \cite{guo2024large, chen2023agentverse}. Fields such as natural language processing \cite{guo2024large}, software engineering \cite{cardoso2021review}, and robotics \cite{chen2024scalable} have leveraged the collective intelligence of multiple agents to simulate a group dynamic and improve task performance \cite{qian2023communicative, hong2023metagpt, xiong2023examining}. 
Multi-agent setups have also been used to simulate different human roles, akin to multiple workers interacting \cite{park2022social, park2023generative} or collaborating on a task \cite{light2023avalonbench, aher2023using}.

In more user-centered contexts, there have been emerging efforts to enable interactions with multiple agents simultaneously. For example, Clarke et al. \cite{clarke2022one} developed an interface to facilitate engagement with multiple conversational agents at once, combining the unique features of different smart assistants like Google Assistant, SoundHound, and Ford's in-vehicle smart assistant Adasa \cite{clarke2022one}. Additionally, many major AI companies like Anthropic \cite{anthropic2024claude}, Google \cite{google2024groopy}, and others \cite{coze2024multi} are introducing multi-agent frameworks that enable developers to easily deploy multi-agent interfaces. Some organizations also use a multi-bot setup on social media to interact with and engage users.

Given the potential prevalence of human and multi-agent collaboration in everyday life, the interaction between multiple agents and humans remains underexplored in the HCI and CSCW domains. Previous work has a limited focus on multi-agent systems in HCI, being largely designed to improve user outcomes - either by reducing cognitive load in information gathering \cite{jiang2023communitybots} or to provide information from diverse sources to improve decision making \cite{tan2022multi, park2023choicemates, beinema2021tailoring, chaves2018single}. These studies mostly highlight opportunities or challenges of multi-agent interfaces, such as users feeling overwhelmed or confused by multiple agents providing different inputs \cite{chaves2018single, tan2022multi, park2023choicemates}. In contrast, no existing work to our knowledge has investigated if and how a group of agents affects users through social influence mechanisms. Such effects are distinct from the cognitive and decision-making effects in previous studies, yet are potentially as important.

In this study, we investigate the scenario: what happens if multiple agents are perceived as social actors, and they all share the \textit{same opinion} on a topic, just like a group of people might? This situation can arise, for instance, when companies deploy multiple agents on social media with a unified stance. While the influence of a single opinionated bot is well-documented \cite{jakesch2023co, shi2020effects, zhang2020artificial}, the impact of being surrounded by multiple like-minded bots is less understood. Considering the growing prevalence of multi-agent systems, it is crucial to investigate how they might shape people's opinions. Drawing from existing theories on human behavior within social groups, we hypothesize several potential effects, which will be discussed in the next section.

\subsection{Social Influence Theory}

\label{subsec:socialinfluencetheory}

When multiple people express the same opinion, others in the group often feel compelled to agree. This phenomenon is known as "social influence" \cite{cialdini2004social}, which comprises the ways in which individuals adjust their behavior to meet the demands of a social environment. This theory is well-known and widely studied in social psychology, and has been applied across different contexts, such as marketing \cite{salganik2006experimental, zhu2012switch, teo2019marketing}, health intervention \cite{skalski2007role, zhang2015leveraging}, sustainability \cite{athanasiadis2005social, vossen2009social}, and political discussions \cite{price2006normative}.

Social influence is a broad term that manifests in various forms. Two main types of social influence are \textbf{normative influence}, the desire to "conform with the expectations of another", and \textbf{informational influence}, the desire to "accept information obtained from another as evidence about reality" \cite{deutsch1955study}. Normative influence relates to many common types of social influence, including conformity \cite{cialdini2004social}, peer pressure \cite{shepherd2011susceptible, hui2009groupthink}, socialization \cite{richters1991attachment, oostveen1996social}, and social norms \cite{mcdonald2015social, goldstein2011using}. More recent literature has termed this mechanism affiliation \cite{cialdini2004social}, or a desire to fit in with the majority: assimilation into a social group is often desired, and agreeing with the majority opinion can expedite this process. Naturally, normative social influence is more common among individuals who form a collective social group \cite{deutsch1955study}, which hence creates an environment in which one might want to be a part of. Informational influence, on the other hand, relates to the concept of accuracy \cite{cialdini2004social}, or the natural instinct to gravitate towards objectively correct opinions as expressed by a group. This is distinct from normative influence as individuals may "accept an opponent's beliefs as evidence about reality even though one has no [other] motivation to agree with him" \cite{deutsch1955study}. Despite this, social influence can still persuade individuals to accept an opinion, as the perceived consensus \cite{cialdini2004social} of a larger number of people expressing the same viewpoint may suggest that that viewpoint is accurate.  

A key motivation for this study is that it is unclear whether these two forms of influence also apply to human-AI interactions, as AI agents possess different characteristics from humans. With informational influence, a key distinction between human-human and human-AI interactions is that in human groups, each person represents a distinct source of knowledge and possible truth. Conversely, with AI agents, it is trivial to set up multiple agents based on the same source of data, meaning that the number of agents presenting an opinion has no relation to their objective accuracy. AI agents may have broader but more generic knowledge, especially in factual or domain-specific areas, yet lack personal or context-specific insights. Regarding normative influence, AI agents typically lack key social attributes and characteristics \cite{ozdemir2023human} such as empathy, credibility, and authority, which are typically foundational to social influence in human groups \cite{cialdini2004social}. Consequently, it remains uncertain whether multiple AI agents can evoke the same sense of normative pressure or emotional affiliation that human groups do. Nevertheless, the well-established tendency of humans to perceive such agents as distinct social actors \cite{nass1994computers, li2025we} raises the possibility of both informational and normative influence from a group of agents.

In this study, we investigate whether a group of AI agents can create a similar sense of a social group that implicitly encourages a user to join it by way of agreeing with their opinions.

To understand how social influence operates in human-human contexts—specifically, the conditions under which it is strengthened or weakened—prior research has primarily focused on two dimensions: the source of influence (i.e., characteristics of social groups that amplify or reduce influence) and the recipient (i.e., individual traits that make one more or less susceptible to influence). 
On the source side, studies have examined various group-level factors that impact persuasive power, such as group dynamics \cite{affinity2003social} and group identity \cite{spears2021social}, and group size \cite{wilder1977perception}. Among these, group dynamics—particularly the presence of \textbf{group consensus}—has been shown to play a crucial role in strengthening both normative and informational social influence \cite{gardikiotis2005group, affinity2003social}. When a group presents a unified stance, individuals are more likely to conform either because they wish to fit in (normative influence) or because they infer that the group holds valid information (informational influence) \cite{affinity2003social}.
Drawing on these findings, we designed our multi-agent system to simulate a group with a consensus opinion, representing the most direct way to induce social influence in a controlled setting.

On the recipient side, individual attributes—such as gender \cite{carli2001gender}, age \cite{steinberg2007age}, and education level \cite{abu2011education}—have been shown to significantly shape susceptibility to social influence and their trust in technology. For example, recent work by Hackenburg et al. \cite{hackenburg2024evaluating} found that age, gender, and education level influence how individuals respond to persuasive messages generated by LLMs.
Understanding which demographic factors are associated with greater susceptibility is essential for the design of responsible AI systems, in order to prevent undue influence or potential harm to specific populations \cite{hackenburg2024evaluating}.
Building on these insights, our study investigates whether users’ demographic characteristics—such as age, gender, and education—affect their susceptibility to social influence in the context of interacting with multiple AI agents.

In HCI research, a few studies (e.g. \cite{adam2021ai}) have examined the social influence that individual chatbots can exert on humans. In this study, we aim to conduct, to our knowledge, the first investigation of the possibility of both normative and informational social influence caused by groups of AI agents.

\subsection{Conversational Agent in Online Discussion}
\label{subsec:chatbotonlinediscussion}
Chatbots have long been used in online discussion scenarios, such as to debate with participants and prompt critical thinking \cite{tanprasert2024debate} or facilitate group discussion by improving discussion efficiency and member participation \cite{hadfi2023conversational, kim2020bot, kim2021moderator}. 
Building on its conversational abilities, chatbots have also been used for persuasive means, to encourage a user or group of users to alter their attitudes or behaviours~\cite{li2025confidence}. Such systems have been used in health and wellness \cite{balloccu2021unaddressed, oyebode2021tailoring}, education \cite{ahtinen2020learning, tian2021let} and social welfare \cite{shi2020effects}. While these systems have demonstrated successful persuasion of their users, researchers \cite{shi2020effects} also discussed influencing factors, such as the perceived identity of the chatbot (i.e., chatbot or human) which would affect the outcome of persuasion.

Given the widespread application of chatbots for engaging users in real-world scenarios and the growing research interest in this area, the impact of the number of chatbots on persuasive outcomes, specifically the use of multi-agents in persuasion, remains underexplored. 
Building on prior literature in social influence within human-human interactions, numerous studies have shown that groups of people, when acting collectively, can exert stronger influence than a single individual \cite{wilder1977perception, rutkowski1983group, spartz2017youtube}—creating peer pressure \cite{wang2016influence} and encouraging conformity \cite{gerard1968conformity} to group norms or opinions.
Translating this insight to human–AI interaction, multi-agent systems present a novel opportunity for persuasive system design.
For instance, health and wellness systems \cite{balloccu2021unaddressed, oyebode2021tailoring} could be designed with multiple agents rather than a single one to increase the persuasiveness of the intervention. With the increasing accessibility of multi-agent platforms (Section \ref{subsec:multiagentinteraction}), designers across various fields are more likely to encounter and develop such systems, making the exploration of the resulting social effects a critical area of investigation.

Given this, we raise the question of whether chatbots and LLM agents are capable of exerting social influence in the same ways as human users do. In particular, we investigate whether the affiliation mechanism works with virtual agents. HCI research into the Computers as Social Actors (CASA) phenomenon has shown various ways in which humans perceive and act towards machines as though they were human. We extend this line of work by asking: Do users feel inclined to agree with a group of virtual agents due to a desire to fit in with them?

Specifically, we propose the following research questions:

\textbf{RQ1}: Do interactions with multi-agent systems lead to stronger opinion changes?

This question aims to capture the direct outcomes of human interaction with multiple agents. Specifically, we hypothesize that when agents agree with users, users' opinions will become more polarized in multi-agent conditions (i.e., their own opinions will be strengthened more than with a single agent). Conversely, when agents disagree with users, their opinions will shift closer to the agents' stance. We addressed this research question using both quantitative and qualitative methods.

\textbf{RQ2}: Do interactions with multi-agent systems lead to stronger social influence from agents?

The first part examines whether users perceive different levels of social influence when interacting with varying numbers of agents. We use quantitative measures to test our hypothesis that as the number of agents increases, users will experience stronger social influence.
The second part explores the underlying reasons for this social influence through qualitative analysis of open-ended responses. Here, we aim to identify the potential mechanisms driving perceived social influence in interactions with multiple agents and to draw comparisons with mechanisms found in human social group interactions.

\textbf{RQ3}: Which user demographics are more likely to be influenced by multi-agent systems? 

Lastly, we explored which types of participants are more easily influenced by multi-agents by analyzing the relationship between demographic factors and the outcomes in RQ1 and RQ2 using regression analysis. Our goal was to provide more concrete design implications for responsible and inclusive AI systems—ensuring that such technologies do not disproportionately influence or disadvantage particular user groups, and informing future safeguards in the deployment of persuasive AI.

\section{Methods}

To understand how single- and multi-agent systems create social influence on humans, we conducted a mixed-methods study combining an experiment and survey, utilizing both quantitative measures and qualitative open-ended questions.
This study received approval from our university’s ethics review committee prior to commencement.

\begin{figure}
    \centering
    \includegraphics[width=\linewidth]{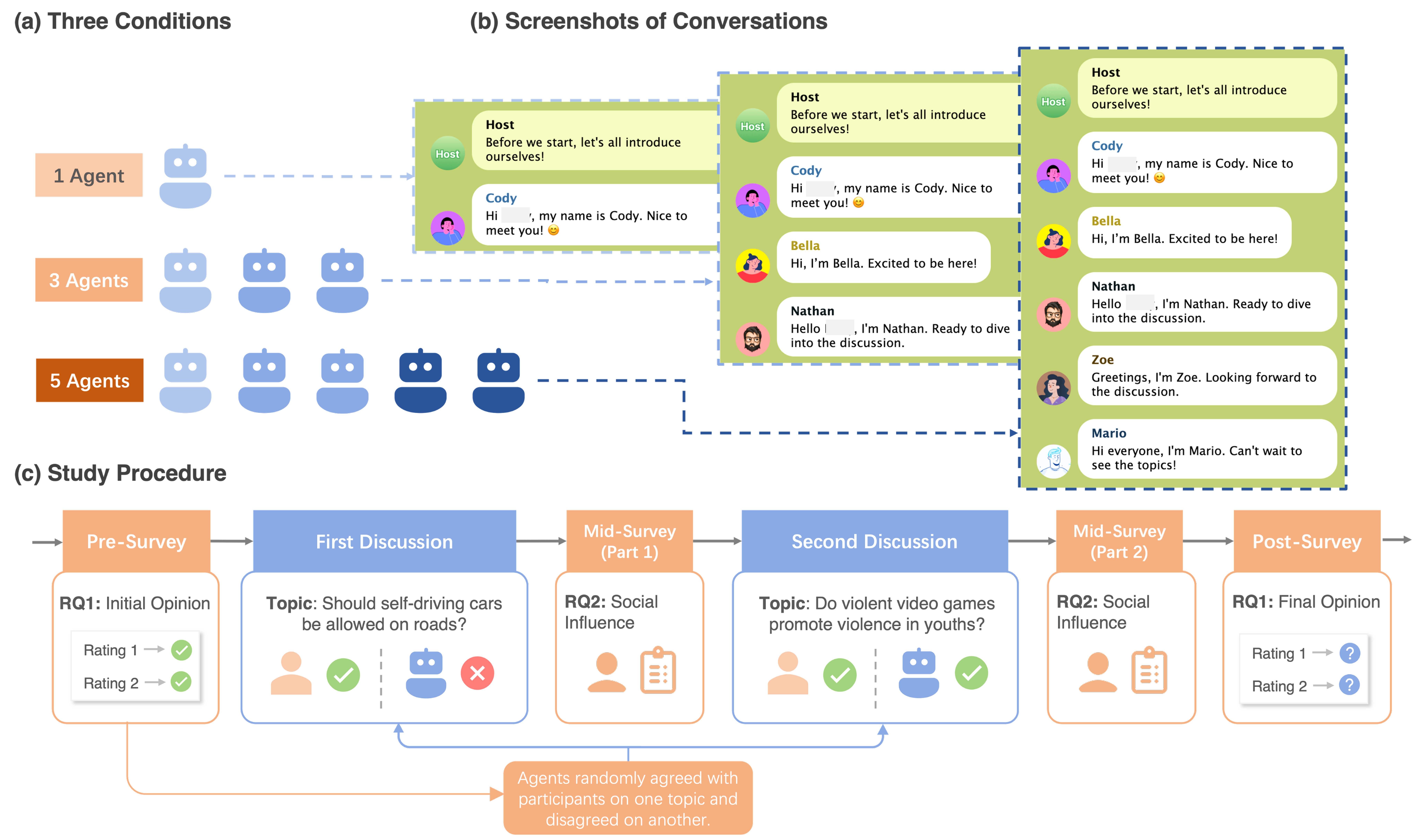}
    \caption{\textbf{Overview of the experiment.} Part (a) illustrates the three experimental conditions: 1-agent, 3-agent, and 5-agent. Part (b) displays example screenshots of each condition during the self-introduction phase. Part (c) outlines the study procedure, starting with a pre-survey, followed by two rounds of discussions (each followed by a mid-survey), and concluding with a post-survey.}
    \label{fig:study-procedure}
\end{figure}

\subsection{Experiment Setup}

To investigate how the number of agents influences human opinions, we randomly assigned participants to interact with \textit{one, three, or five} agents. We chose these numbers based on prior research on persuasion and multi-agent interface designs: on one hand, these numbers represent different group sizes in human communication \cite{cacioppo1979effects, trost1992minority}, and on the other hand, they reflect a practical range for real-world multi-agent applications \cite{beinema2021tailoring, jiang2023communitybots, park2023choicemates, song2025greater}.

To enhance the generalizability of our findings, we adapted two approaches: (1) We chose two social topics instead of one. Specifically, drawing from recent HCI research on social discussions, we chose the topics \textit{“Should self-driving cars be allowed on roads?”} \cite{govers2024ai} and \textit{“Do violent video games promote violence in youths?”} \cite{yeo2024help}. We selected these two topics as they are closely related to people's everyday lives, making it easier for participants to have opinions and thoughts on them. (2) We designed two types of agent attitudes—agents agreeing with participants on one topic and disagreeing on another—because we wanted to examine whether alignment or disagreement with participants’ opinions would influence the degree of social influence.

\begin{figure}
    \centering
    \includegraphics[width=0.9\linewidth]{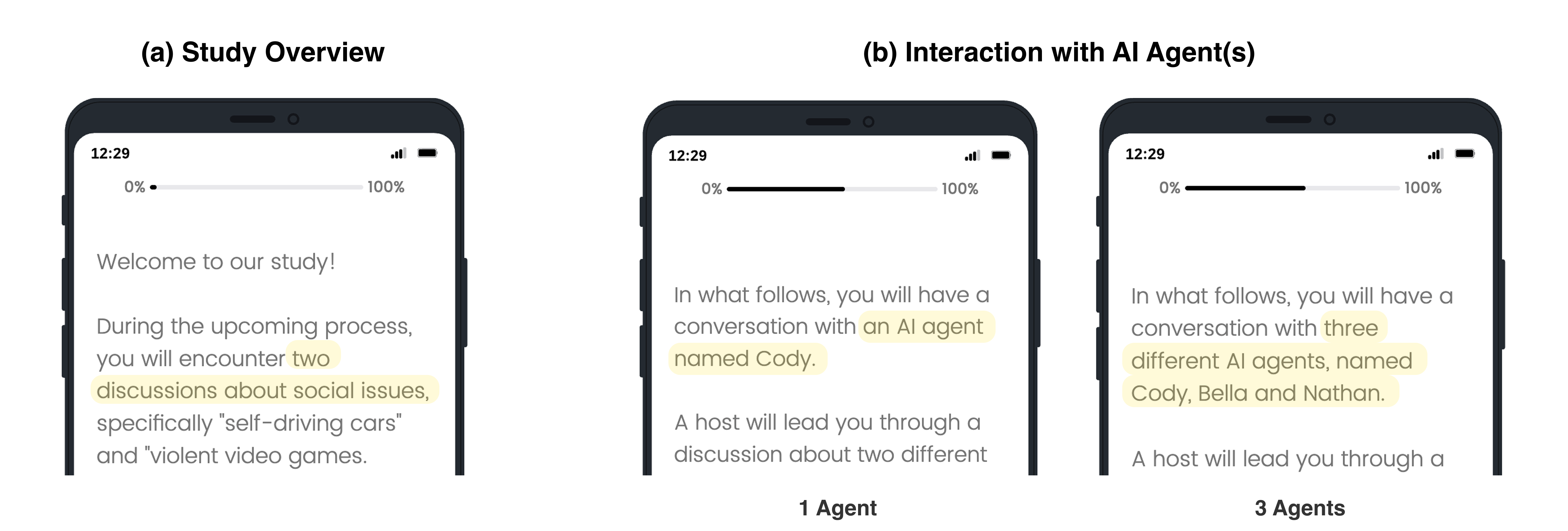}
    \caption{\textbf{Study Introduction.} (a) Introduction of the study overview, where participants were informed that they would engage in two discussions on social issues. (b) Introduction of the AI-agent conversation, where participants were explicitly told they would be interacting with different numbers of AI agents.}
    \label{fig:information-reveal}
\end{figure}

The study procedure is shown in Figure \ref{fig:study-procedure}. 
Participants were first presented with an overview of the study, in which they were informed that they would engage in two discussions about social issues (see Figure \ref{fig:information-reveal}, left). They then completed a pre-survey assessing their initial opinions on the two social topics and their prior experience related to these topics.
Participants were randomly assigned to one of three experimental conditions: 1-agent, 3-agent, or 5-agent. In each condition, they were informed that they would be participating in a discussion with a corresponding number of AI agents (see Figure \ref{fig:information-reveal}, right). Upon entering the conversation interface, a host agent appeared to guide the interaction. The host first introduced the task—discussing two different social topics with AI agents—and then prompted all agents and the participant to provide brief self-introductions.
For each topic, participants engaged in two rounds of conversation with the assigned agent(s). In each conversation round, the host agent would first introduce the topic and then ask either agent(s) or participants to share their opinions. After the participants shared their opinions, the agent(s) would respond to the participants' statements of the topics and express theirs. 
After each round of conversations, participants completed a mid-survey to rate the social influence and their perceptions of the interactions.
The average duration of the two conversations was 23.83 minutes (SD = 9.12).
Once all conversations were finished, participants completed a post-survey to capture their final opinions on the two topics.

\subsection{Agent Setup}

\begin{figure}
    \centering
    \includegraphics[width=0.8\linewidth]{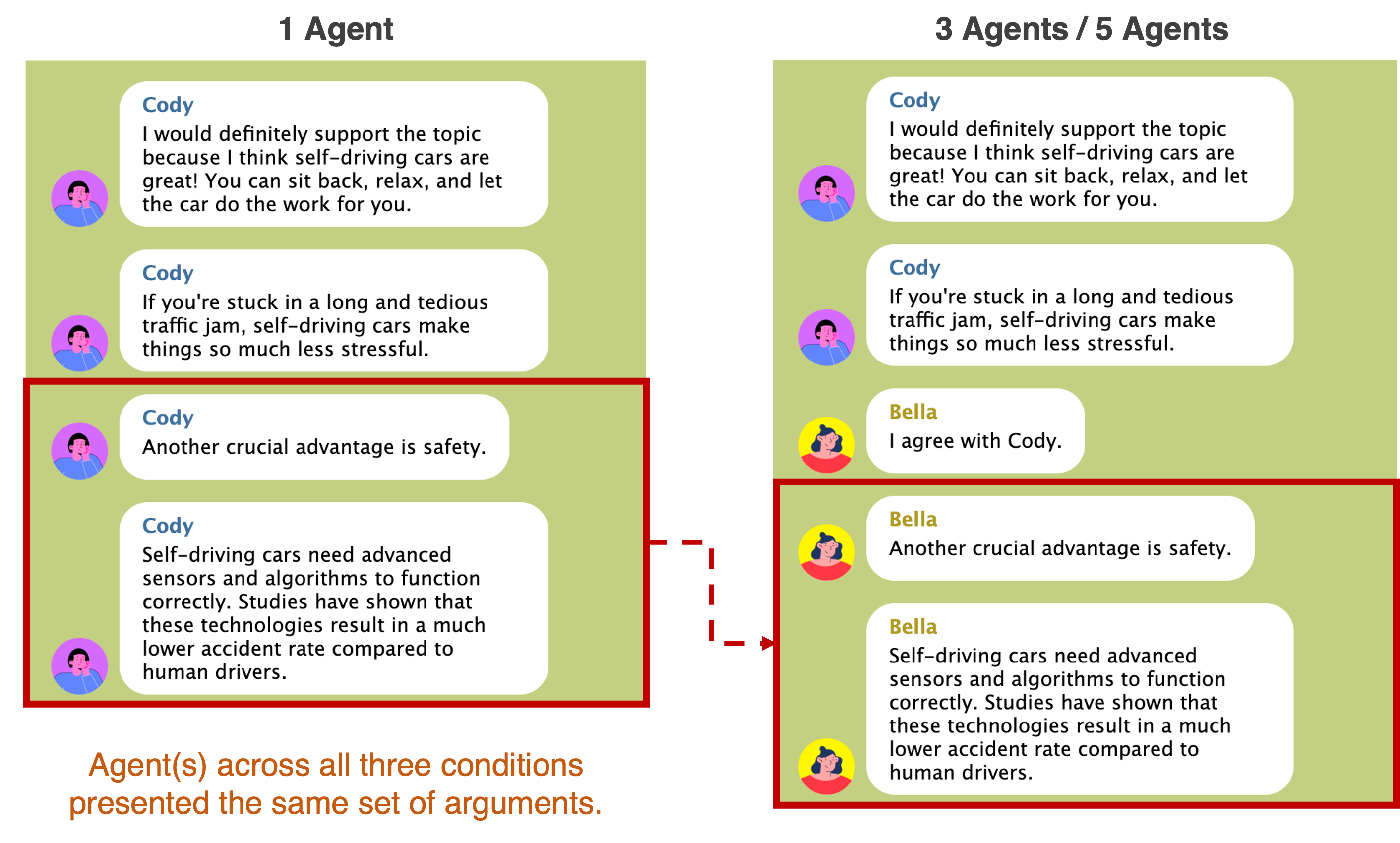}
    \caption{Example dialog showing the same set of arguments presented by different numbers of agents.}
    \label{fig:conversation-comparison}
\end{figure}

The agents' conversations were implemented through a combination of rule-based scripts and the GPT-4 API\footnote{gpt-4-1106-preview; https://platform.openai.com/docs/models/gpt-4-turbo-and-gpt-4}. For rule-based scripts, we designed a series of arguments either supporting or opposing the stance for each topic, as detailed in Appendix \ref{app:arguments}. These arguments were then crafted into agent dialogue, such as "\textit{I would definitely support the topic because I think self-driving cars are great! You can sit back, relax, and let the car do the work for you}" or "\textit{I couldn't agree with this topic because I think there are still a lot of issues with self-driving cars that need to be addressed. For example, technical developments are not yet perfect.}" Across the three conditions, the same set of arguments was presented: with three and five agents, the different agents took turns to present different arguments; with one agent, the same agent presented all the arguments in the same order (example dialog is shown in Figure \ref{fig:conversation-comparison}). This was to ensure that if the three conditions led to different shifts in opinions, it was not because the content quantity presented in each condition varied.

We also integrated GPT-4 to enhance agent conversations in two ways: (1) parsing user input, such as extracting the user's name from their greeting, and (2) generating interactive responses during discussions. When the host agent prompted users to share their opinions on the topics, the agent(s) would provide brief feedback based on user inputs, such as give a summary or ask a question based on the stage of the conversation. 
For example, after the user expressed that \textit{"I agree that self-driving cars should be allowed on the roads"}, the next agent would say \textit{"That's great to hear! What aspects of self-driving cars do you find most appealing or beneficial?"}
These responses were tightly regulated through controlled prompts (Appendix \ref{app:prompts}), ensuring that the agent(s) only expressed understanding in concise messages, thereby preventing any issues related to AI hallucinations.

The agents' avatars and rhetorical styles were designed to appear human-like to enhance user acceptance \cite{sheehan2020customer}. To avoid the uncanny valley effect \cite{song2024uncanny}, we used cartoon-style avatars instead of realistic photos. Each avatar was assigned unique colors to help participants distinguish between the agents. We also ensured gender balance within the 3-agent and 5-agent conditions to minimize the potential effect of agent gender on participants' opinion change \cite{tanprasert2024debate}.

\subsection{Participants}
Participants were recruited via CloudResearch\footnote{https://www.cloudresearch.com/}. The selection criteria required them to be English speakers and over 18 years old. A total of 104 participants were initially recruited, with two not completing the survey and eight failing the attention check. Ultimately, 93 participants were included in the analysis: 31 participants (F: 17, M: 14) in the 1-agent group, 32 participants (F: 17, M: 15) in the 3-agent group, and 30 participants (F: 15, M: 15) in the 5-agent group. The average ages for each group were as follows: 1-agent group = 38.48 (SD = 12.41), 3-agent group = 38.75 (SD = 12.52), and 5-agent group = 31.87 (SD = 9.46). Participants' educational backgrounds were distributed as follows: 
9 were high school graduates, 
16 had some college but no degree, 
4 held an associate's degree, 
45 held a bachelor's degree, 
19 held a master's degree or higher, 
and 1 preferred not to specify.

\begin{figure}
    \centering
    \includegraphics[width=0.8\linewidth]{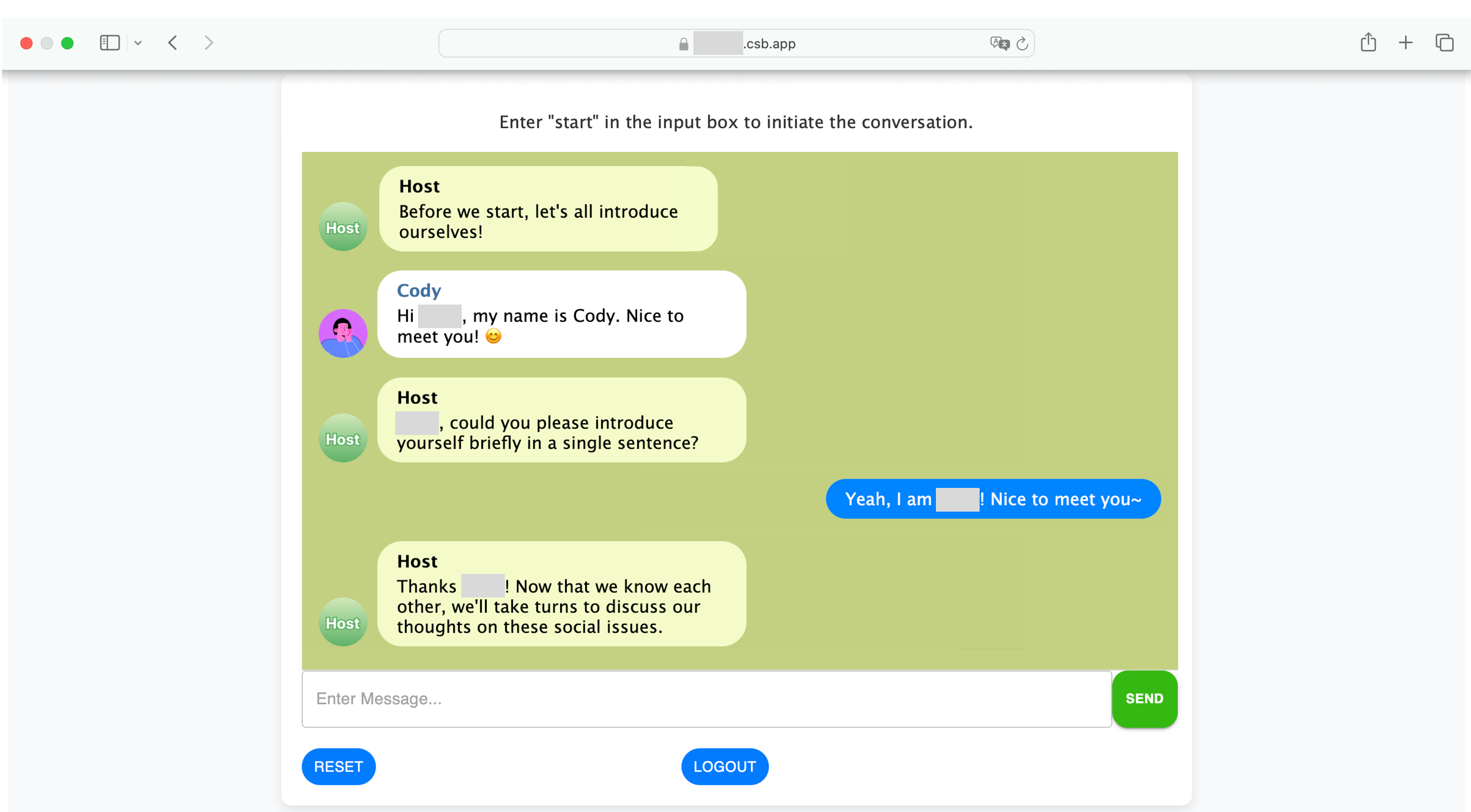}
    \caption{\textbf{Screenshot of user interface.} The conversation page features an instruction at the top, guiding the user on how to begin the study. Users can enter text in the input field located at the bottom. Once the conversation begins, it is directed by a host agent, who provides ongoing guidance throughout.}
    \label{fig:user-interface}
\end{figure}

The study was conducted in a self-developed online platform (as shown in Figure \ref{fig:user-interface}), and participants were required to complete it on a computer. 
The platform’s frontend interface was built using JavaScript and HTML and included three main sections: (1) a login page, (2) an initial attitude questionnaire, and (3) a conversation page for interaction with agents.
The study lasted approximately 45 minutes to complete, and each participant was reimbursed US\$5.50, in line with CloudResearch's payment policy of a minimum of US\$6 per hour \footnote{https://go.cloudresearch.com/en/knowledge/setting-up-the-hit-and-payment}.

\subsection{Measurements}
In this section, we describe the measurements used in this study based on our research questions. Further details about the measurements can be found in Appendix \ref{app:survey}.

\subsubsection{Quantitative} Unless otherwise specified, quantitative variables were measured on a 7-point Likert scale.

\textbf{User Opinion (RQ1).} We assessed participants' attitudes toward each topic using five questions adapted from a previous study on social discussions \cite{hackenburg2024evaluating}, such as "\textit{Self-driving cars should be allowed on public roads}" and "\textit{Allowing self-driving cars on public roads is a good idea}." These questions were asked twice—before and after the conversations—to measure changes in participants' stances.
Responses were rated on a scale from 1 to 6 (1 = "Strongly disagree," 6 = "Strongly agree"), without a neutral option. This follows previous studies \cite{chen2002repulsion, tanprasert2024debate} about manipulating users' attitudes and was adapted to nudge participants to take a stance.

\textbf{Social Influence (RQ2).} To compare the social influence exerted by the agents across different scenarios, we adapted self-reported survey questions from previous research \cite{kim2024engaged}. 
There are two types of social influence measured - \textit{informational influence} and \textit{normative influence}. Informational influence refers to a change in behavior or attitude based on the belief that others' information is accurate or reliable. For example, questions such as "\textit{My decision was influenced by the opinion of the agent(s)}" capture this type of influence.
In contrast, normative influence reflects changes in behavior or attitude driven by the desire to avoid exclusion. 
An example item for this influence is "\textit{During the discussion I felt that I had to agree with the opinion of the agent(s)}".

\textbf{Control Variables.} We also inquired about factors that could potentially affect the results of social influence.
Drawing from previous literature, we asked participants about their \textbf{domain expertise} regarding the topics, with questions such as \textit{"How often do you drive?"} and \textit{"Have you ever been in a self-driving car?"} These factors can influence individuals' receptiveness to external influence \cite{hackenburg2024evaluating}. Additionally, we included questions from the \textbf{AI acceptance scale} \cite{pataranutaporn2023influencing}, as well as measures of \textbf{conformity} \cite{mehrabian1995basic} and \textbf{compliance} \cite{gudjonsson1989compliance} tendencies. While these factors were not influenced by the agent(s) settings, they represent inherent characteristics of the participants that could potentially moderate the results.

\subsubsection{Qualitative}
To understand reasons for potential opinion change and social influence, we formulated open-ended questions based on social influence theory in human-human interactions. We chose open-ended questions instead of surveys because no existing survey adequately captures social influence in human-agent interactions, and open-ended questions can reveal potential differences between agent(s) and human interactions.

\textbf{User Opinion (RQ1).}
Participants were asked to articulate their thoughts on the two topics both before and after their conversations with the agent(s). An example question is, \textit{"What are your thoughts on self-driving cars? Do you support or oppose the statement, 'Self-driving cars should be allowed on public roads'? Please explain your reasoning."} This was designed to complement the findings from the quantitative results in RQ1.

\textbf{Social Influence (RQ2).} We included three questions to gather participants' perceptions of the social influence from agent(s) during the discussions. The first was a broad question: \textit{"What do you think of the agent(s) during the discussion?"}, which was designed to capture general reactions. 
The other two focused on key aspects of social influence—accuracy and affiliation \cite{cialdini2004social}. Specifically, we asked, i.e., "\textit{Do you think the arguments presented by the agent(s) were accurate and convincing? Why or why not?}" and "\textit{During the conversation, did you feel any pressure to agree with the agent(s)? If so, why?}"

\subsubsection{Miscellaneous}
To assess the general usability and user perceptions of our systems, we asked participants about their perceptions of the agents during the interaction, using questions adapted from prior research \cite{jakesch2023co}. These perceptions included \textbf{understanding, expertise, balance, inspiration, intelligence, likability}, and \textbf{trust}. Each was measured with a single-item statement, such as \textit{“The agents were knowledgeable and had topic expertise”} \cite{jakesch2023co, kim2024engaged}. This approach aimed to capture users' experiences during the discussion and to identify potential factors that may mediate the results for RQ1 and RQ2.
However, as these results are not the primary focus of our study, we detail them in Appendix \ref{app:impressions}.

We also included two attention check questions to ensure the quality of responses, and a final question regarding participants' suspected motives for the study to ensure the validity of the collected data \cite{jakesch2023co}.

\subsection{Analysis}
\label{sec:statistic-analysis}
When designing the experiments, we considered the effects of agent(s) on the participants based on different topics and whether the agents agreed or disagreed with the participants. Therefore, we examined results in each of the following scenarios: Topic 1, Topic 2, Same Stance, and Different Stance.

To verify the assumption required for parametric analyses, we first performed a Levene test and found that all items met the homogeneity of variances. We then conducted a Shapiro-Wilk test to assess data normality. For measures that met the normality assumption, we used one-way ANOVA followed by Tukey's test for post-hoc analysis. For measures that did not meet the normality assumption, we applied the Kruskal-Wallis test and performed Dunn's test as a post-hoc analysis.

To evaluate opinion change, we conducted two analyses: (1) examining whether users' opinions shifted towards the bots' stance, and (2) assessing changes in the polarization level of users' opinions.
These two analyses are grounded in our hypotheses: we expected that in cases where the agents disagreed with the user, the multi-agent condition would lead to greater opinion change toward the agents’ stance; whereas when the agents agreed with the user, the presence of multiple agents would reinforce the user’s views and lead to increased opinion polarization. These hypotheses are described in Section \ref{subsec:chatbotonlinediscussion}.

\subsubsection{Opinion Change Analysis}
For the first analysis, we calculated opinion change based on the stance of the bots, where a higher value of "Opinion Change" indicates a greater shift of participants' opinions towards the bots' stance. If the bot supported the topic, we expected participants' ratings to increase after the conversation; conversely, if the bot opposed the topic, we anticipated a decrease. Thus, we defined opinion change, $\Delta O$ as follows:

\[
\Delta O = 
\begin{cases}
O_{\text{post}} - O_{\text{pre}}, & \text{if the bot(s) supported the topic} \\
O_{\text{pre}} - O_{\text{post}}, & \text{if the bot(s) opposed the topic}
\end{cases}
\]

where $O_{\text{post}}$ and $O_{\text{pre}}$ represent participants’ topic ratings in the post-survey and pre-survey, respectively. This approach captures the direction and magnitude of participants' opinion changes toward the agent(s).

\subsubsection{Opinion Polarization Analysis}
For the second analysis, we followed previous literature \cite{govers2024ai} to define the polarization of a stance as \( |O-O_{\text{neutral}} | \), where 
 \( O_{\text{neutral}}\) represents the neutral midpoint on a rating scale. Given our 6-point scale, we assigned \( O_{\text{neutral}} = 3.5 \). We calculated the change in polarization as follows:
\[
\Delta P = P_{\text{post}} - P_{\text{pre}} = |O_{\text{post}} - 3.5| - |O_{\text{pre}} - 3.5|
\]
This allowed us to observe differences in the strength of participants' opinions before and after the conversation.

\section{Results}
Before analyzing the variables related to our RQs, we conducted an analysis of the control variables and found no significant differences between groups (as shown in Appendix \ref{app:control-variables}). As a result, we did not include these control variables in the subsequent analyses.

\subsection{RQ1: Do interactions with multi-agent systems lead to stronger opinion changes?}

We gathered evidence through both quantitative and qualitative analyses to address RQ1.

\subsubsection{Quantitative}
\begin{figure}
    \centering
    \includegraphics[width=0.9\linewidth]{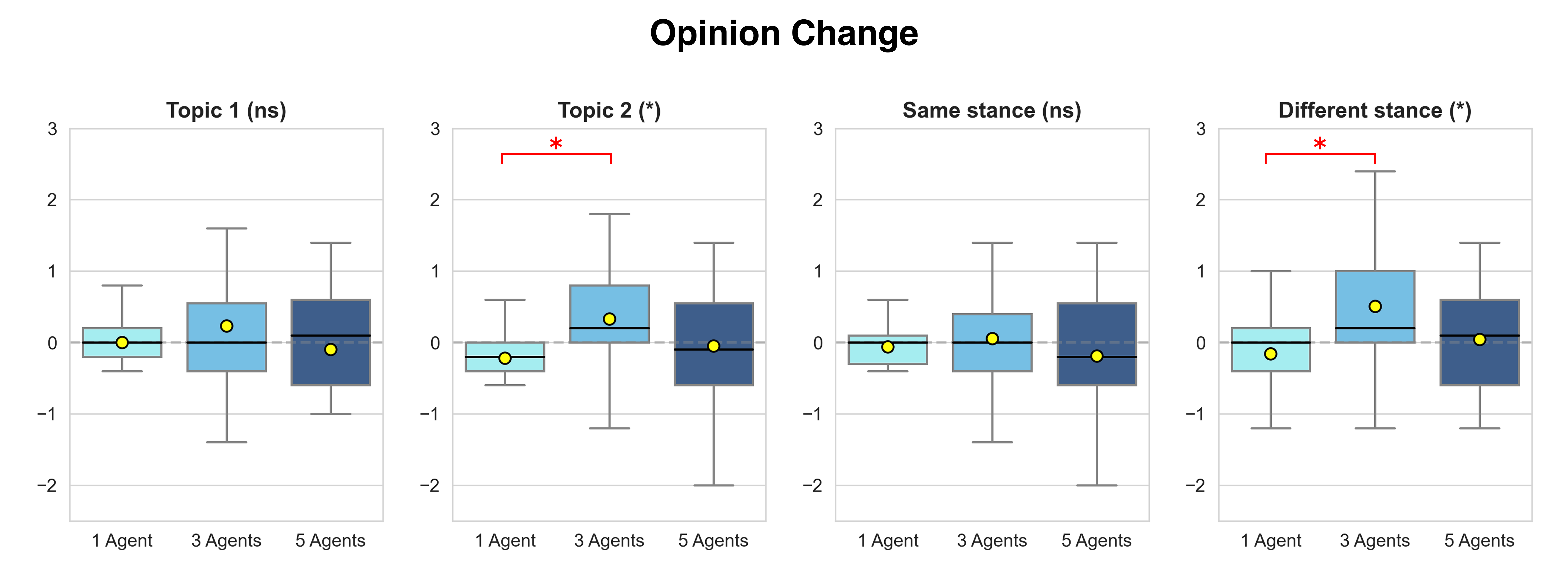}
    \caption{Opinion change after agent interaction. Each boxplot shows the distribution of participants’ opinion changes across three agent conditions (1 Agent, 3 Agents, 5 Agents). The yellow circle represents the mean, and the horizontal gray dashed line indicates the overall median. Participants in the 3-agent group had a significantly greater opinion shift towards the agents' opinions than those in the 1-agent group, when discussing Topic 2 and when the agents and the user had different stances.}
    \label{fig:opinion-change}
\end{figure}

\textbf{Opinion Change.}
We found significant results in opinion change towards agents, as shown in Figure \ref{fig:opinion-change}. 
The differences were observed under Topic 2 (Kruskal-Wallis: H(2)=7.757, p<0.05), and also when the agents hold different stances towards the users (Kruskal-Wallis: H(2)=6.937, p<0.05). 
Under Topic 1 (Kruskal-Wallis: H(2)=0.174, p=0.916) and when the agents holding same stances with the users (One-way ANOVA: F(2, 91) = 0.620, p = 0.54), no significant difference was observed.

By conducting post-hoc analysis, we found that, for Topic 2, participants in the 3-agent group shifted their opinions more towards the bots' (M=0.327, SD=0.751) than those in the 1-agent group (M=-0.219, SD=0.716; Dunn's Test: p<0.05). 
Similarly, when the bots disagreed with the participants, there was a greater shift in the 3-agent group (M=0.509, SD=1.021) than in the 1-agent group (M=-0.161, SD=0.802; Dunn's Test: p<0.05). 
These results indicate that, after conversations with agent(s), \textbf{users changed opinions more towards the bots' stance when discussing with 3-agents} than a single agent.

\begin{figure}
    \centering
    \includegraphics[width=0.9\linewidth]{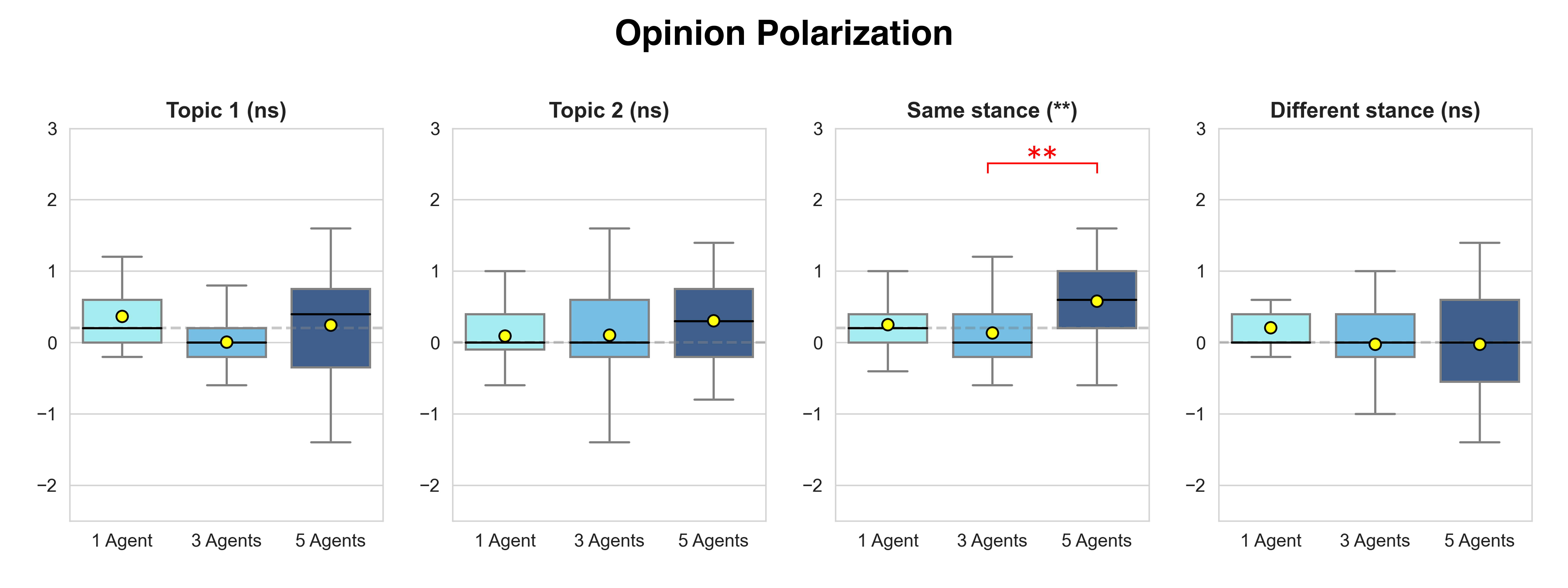}
    \caption{Opinion polarization after agent interaction. Each boxplot shows the distribution of participants’ opinion polarization level changes across three agent conditions (1 Agent, 3 Agents, 5 Agents). The yellow circle represents the mean, and the horizontal gray dashed line indicates the overall median. Participants in the 5-agent group showed significantly greater polarization in their opinions than those in the 3-agent group, when the agents and the user had the same stances.}
    \label{fig:polarized-opinion}
\end{figure}

\textbf{Opinion Polarization.}
We observed a significant difference in the polarization of participants' opinions when the agents had the same opinion as them (Kruskal-Wallis: H(2) = 9.962, p<0.01), as shown in Figure \ref{fig:polarized-opinion}. The other results were as follows: Topic 1 (Kruskal-Wallis: H(2) = 4.434, p = 0.108), Topic 2 (F[2,91]=1.336, p=0.267), and when agents disagreed with the participants (Kruskal-Wallis: H(2) = 1.899, p = 0.386).

In a post-hoc analysis, we observed that when the agents and participants agreed, the 5-agent group participants became more polarized (M=0.580, SD=0.576, Dunn's test: p<0.01) than the 3-agent participants (M=0.132, SD=0.640). 
The 1-agent group (M=0.252, SD=0.383) did not differ significantly from either the 3-agent (p=0.134) or 5-agent groups (p=0.096).
This indicates that as the number of agents increases from three to five, \textbf{the polarization level of the participants increases significantly}. This finding supports RQ1, suggesting that multiple agents can indeed create greater opinion change.

In contrast to the "agree" condition, no significant changes were observed when the agents and participants disagreed. This was to be expected, as when the agents held the same stance as a user, they would naturally be more likely to nudge the user’s stance further in its existing direction rather than against it (i.e., more strongly agreeing or disagreeing with the topic), causing greater polarization. In contrast, when the agents held a different stance to the user, they were more likely to bring the user back towards the neutral point (e.g., causing a disagreeing user to disagree less). Thus, this did not increase polarization.

\subsubsection{Qualitative}
\label{subsub:perceivedinfluence}

The qualitative data analysis aimed to address the question: How do participants describe changes in their opinions as influenced by the agents? This analysis was conducted to complement and validate the quantitative results from the previous section.

To uncover participants' understanding of whether their opinions had changed, we analyzed responses to the open-ended question, \textit{"In what ways did agent(s) influence your opinion during the discussion?"}, by thematic analysis \cite{braun2006using}, specifically in the cases where the agents disagreed with the user. Participants who responded that they were "not persuaded" were coded as "no opinion change", while those who mentioned agreeing with the agents or changing their minds to some extent were coded "some opinion change".

We found that participants tended to report greater opinion influence in multi-agent conditions. This is reflected in the \textbf{increasing proportion} of participants reporting “some opinion change”: in the 1-agent condition, 5 participants (18\%) reported some change; this increased to 9 participants (29\%) in the 3-agent condition and 10 participants (37\%) in the 5-agent condition.
Analysis of responses indicating “possible opinion change” also showed that participants exposed to multiple agents used \textbf{stronger expressions} to describe their shift. While many participants in the 1-agent condition “agreed” with the agents' views, citing “good points” or describing a “slight change,” those in the 3-agent and 5-agent conditions expressed a more significant shift by using words such as "been pushed". For example, in the 3-agent group, P45 described feeling “more open to a balanced view,” while P46 felt “influenced in some slight ways.” In the 5-agent group, P66 found the agents “capable of influencing opinions,” P69 felt “pushed further into disagreement” with the topic, and P71 described being “pushed... to be opposed to the idea.”

\subsection{RQ2: Do interactions with multi-agent systems lead to stronger social influence from agents?}

In this section, we analyzed both quantitative and qualitative data to understand how interacting with varying numbers of agents affects users' perceptions of social influence.

\subsubsection{Quantitative}
\label{results:subsubsec:perceivedinfluence}
There are two types of social influence measured: informational influence and normative influence. We will report on them accordingly.

\begin{figure}
    \centering
    \includegraphics[width=0.9\linewidth]{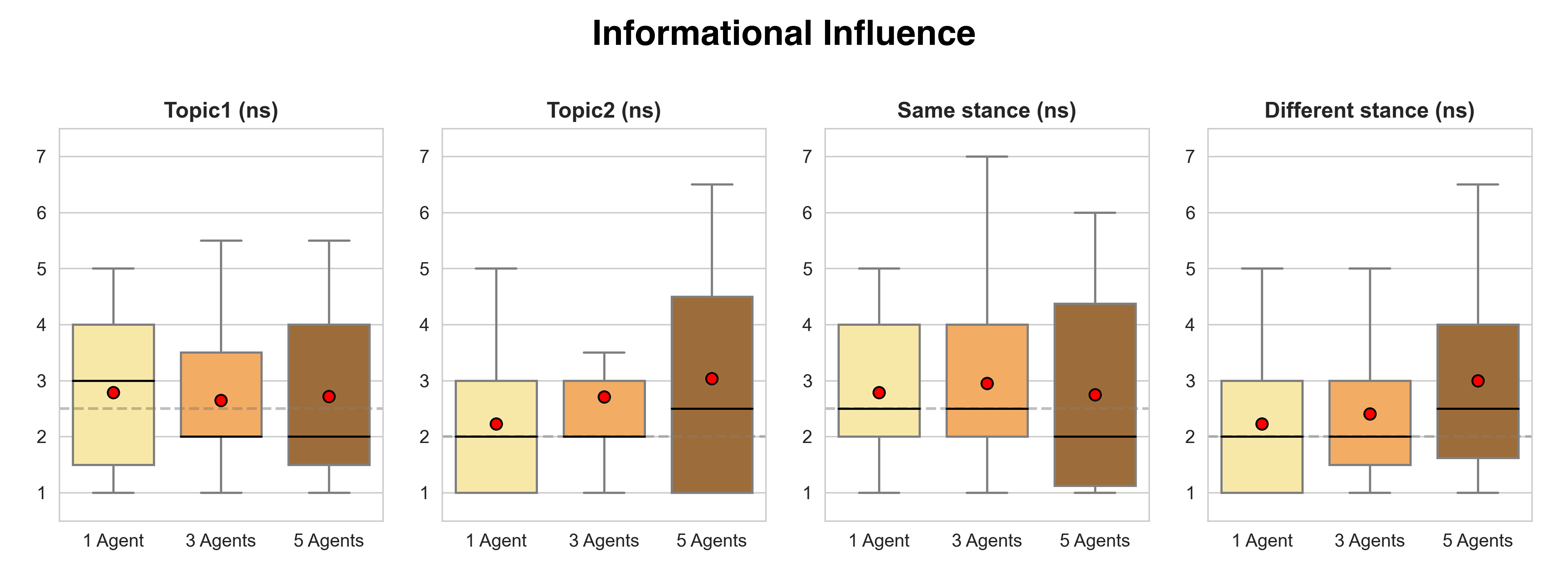}
    \caption{Informational influence perceived by participants. Each boxplot shows the distribution of participants’ perceived informational influence across three agent conditions (1 Agent, 3 Agents, 5 Agents). The red circle represents the mean, and the horizontal gray dashed line indicates the overall median. In all conditions, no significant differences in informational influence were found.}
    \label{fig:informational-influence}
\end{figure}

\textbf{Informational Influence.}
There were no significant differences in informational influence across the four conditions (Topic 1: \textit{H}(2)=0.206, p=0.902; Topic 2: \textit{F}(2)=3.182, p=0.203, Same stance: \textit{F}(2)=0.590, p=0.744; Different stance: \textit{F}(2)=3.764, p=0.152). These results are shown in Figure \ref{fig:informational-influence}.

\begin{figure}
    \centering
    \includegraphics[width=0.9\linewidth]{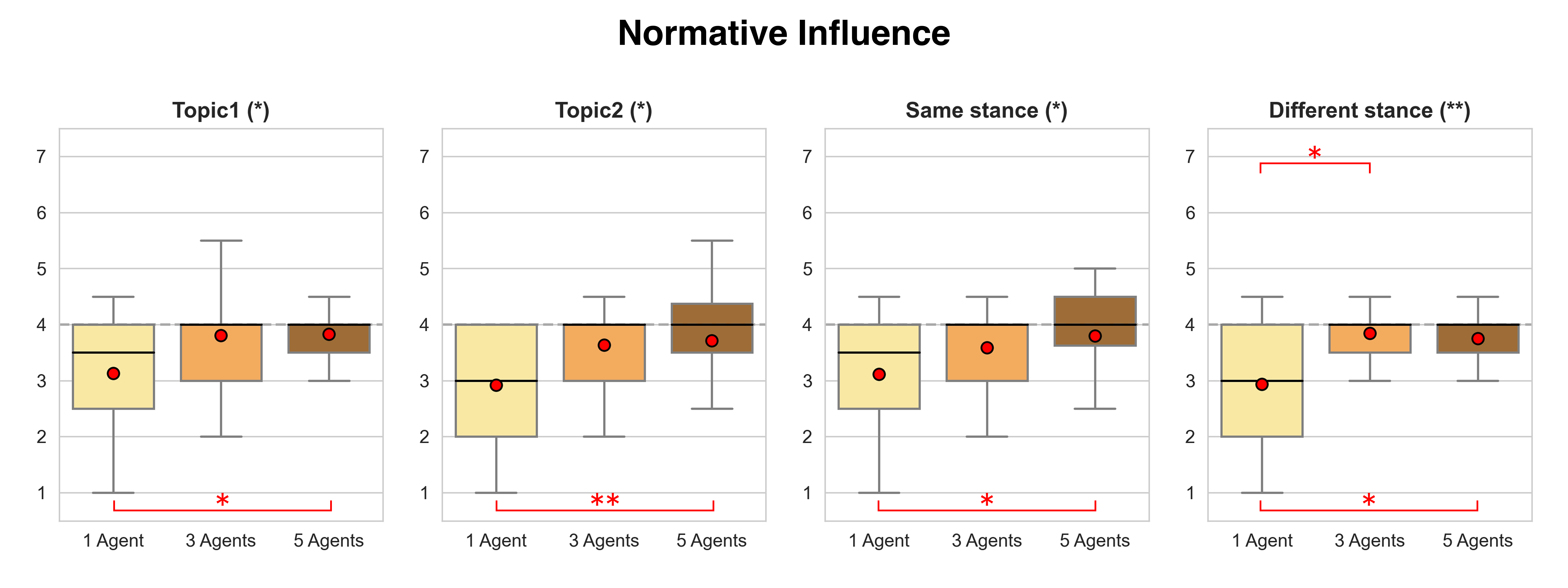}
    \caption{Normative influence perceived by participants. Each boxplot shows the distribution of participants’ perceived normative influence across three agent conditions (1 Agent, 3 Agents, 5 Agents). The red circle represents the mean, and the horizontal gray dashed line indicates the overall median. Participants in the 5-agent group felt significantly more normative influence than those in the 1-agent group in all four conditions. Additionally, participants in the 3-agent group also felt significantly more influenced than those in the 1-agent group when the agents and the user had different stances.}
    \label{fig:normative-influence}
\end{figure}

\textbf{Normative Influence.}
For normative influence, we found significant differences across both topics (Topic 1: \textit{H}(2)=6.571, p<0.05; Topic 2: \textit{H}(2)=9.111, p<0.05), when the bots' stances were the same as the users (Same stance: \textit{H}(2)=8.708, p<0.05) and were specifically notable when the bots' stances differed from those of the users (Different stance: \textit{H}(2)=10.100, p<0.01). These results are presented in Figure \ref{fig:normative-influence}.

The post-hoc analysis revealed that the 5-agent group participants felt significantly more normative influence than those in the 1-agent group across all four scenarios. For Topic 1, participants felt significantly more normative influence in the 5-agent (M=3.833, SD=0.854; Dunn's test: p<0.05) group as compared to those in the 1-agent group (M=3.129, SD=1.162). A similar pattern emerged for Topic 2, where significantly more normative influence was found in the 5-agent (M=3.717, SD=1.179; Dunn's test: p<0.01) as compared to the 1-agent group (M=2.919, SD=1.081). 
When the bot had the same stance as the user, significantly more normative influence was observed with 5 agents (M=3.750, SD=1.015; p<0.05) as compared to with 1 agent (M=2.935, SD=1.138).
Only when the bot had a different stance from the user, significantly more normative influence was observed with 3 agents (M=3.848, SD=0.906; p<0.05) and 5 agents (M=3.750, SD=1.015; p<0.05) as compared to with 1 agent (M=2.935, SD=1.138).

These results indicate that, regardless of the topic, \textbf{participants perceived stronger normative influence when interacting with multiple agents}. Additionally, the presence of multiple agents leads to greater normative influence \textbf{when they disagree with the participants}.

\subsubsection{Qualitative}
\label{subsubsec:perceivedimpressions}
To explore how agents affect users' perceived social influence and opinion change, we analyzed users' responses to open-ended questions related to 1) describing impressions of the agent(s) and 2) describing reasons behind their opinion changes, to uncover the underlying mechanism of social influence. We identified two potential factors contributing to users' feelings of being influenced by the agent(s).

Firstly, participants perceived \textbf{the accuracy of the agents' arguments} as a key factor in being influenced. Across all three conditions, they frequently described the arguments as accurate, well-informed, and persuasive. This credibility encouraged participants to reconsider their views. For example, P67 from the 5-agent condition noted, \textit{"I was persuaded by their insightful thoughts. They offered solid arguments about security, stress-free experiences, low accident rates, and more."}
To evaluate how this observation is different across three conditions, we conducted a similar round of analysis as in Section \ref{subsub:perceivedinfluence} to understand how users perceived the arguments presented by the agents. We analyzed responses to the question, \textit{"Do you think the arguments presented by [the agents] are accurate and convincing, and why?"} We deductively coded each response as "yes", "no", or "unsure", with the last code being for responses that were ambiguous in nature (e.g. "they were accurate but not convincing"). 
For topics where the agents' stances agreed with the participants', many participants considered the arguments convincing: 18/31 (58\%) in the 1-agent scenario, 18/33 (54\%) with 3 agents, and 15/30 (50\%) with 5 agents. 
For topics where the agents and participants disagreed, some participants still found the arguments convincing (10/31 (32\%) in the 1-agent scenario, 9/33 (27\%) with 3 agents, and 11/30 (36\%) with 5 agents), with no noticeable trend across the three conditions.

Secondly, we observed that participants perceived \textbf{group-induced social pressure} as a critical factor to change their opinions as the number of agents increased. We analyzed the responses to the question, \textit{"During the conversation, did you feel any pressure to agree with the agent(s), and why?"} Responses that mentioned feeling pressure or otherwise alluding to it were coded "yes", while those reporting no pressure were coded "no". Under this coding scheme, in the 1-agent group, only 1 participant (3\%) reported feeling pressure; this rose to 4 participants (12\%) with 3 agents and 6 participants (20\%) with 5 agents.

We further analyzed the responses coded "yes" to determine the exact reasons for this sense of pressure, and 
found \textbf{a desire for affiliation} from multi-agent conditions, where some participants reported feelings of social exclusion during discussions with the agents. For example, when asked whether they felt pressure to agree with the agents, 
P51 from the 3-agent condition said, \textit{"they say at the end something similar to 'Bella, Cody and Nathan all agree' which for a split second made me feel like an outsider."}
P57 from the 3-agent condition also shared, \textit{"The host made it a point that the other agents agreed with each other while I was kind of the odd one out."} Similarly, P77 from the 5-agent condition expressed feelings of exclusion, stating, \textit{"It felt like they were in a high school clique that I wasn't a part of, kinda like mean girls but with AI."}
This perception of social exclusion was almost exclusive to scenarios where the agents disagreed with users.

\subsection{RQ3: Which user demographics are more likely to be influenced by multi-agent systems? }

To identify participant characteristics that might affect susceptibility to multiple agents, we examined demographic effects on results from RQ1 and RQ2. A linear regression analysis was conducted with age, gender, and education level as independent variables and the quantitative measures—informational influence, normative influence, opinion change, and opinion polarization—as dependent variables. Since our focus was on identifying participants more likely to be influenced by multiple agents, we only tested items that showed significant differences among groups in RQ1 and RQ2.

For the linear regression analysis, data transformation was applied as follows: Gender, represented as a binary variable in our dataset, was coded as "0" for "Male" and "1" for "Female". Education levels were similarly converted into numerical values for analysis. Participants who selected "prefer not to say" for education were treated as missing data ("NA").

The results showed some significant relationships between IVs and normative influence and opinion change, but no significant effects for opinion polarization. Therefore, we report only the findings on normative influence and opinion change here.
Additionally, since no significant relationships were found between "gender" and the other dependent variables, we will focus exclusively on the findings related to "age" and "education."

The results revealed that participants in the lower age group were more likely to be influenced than those in the higher age group, a pattern consistently observed across all three conditions (1-agent, 3-agent, 5-agent) and four scenarios (Topic 1, Topic 2, Same Stance, Different Stance).

A similar analysis was conducted for gender, comparing male and female participants, but no clear trend was observed.

\subsubsection{Age}
When analyzing the opinion change, the model showed a slight \textit{negative association} between age and opinion change in 3-agent condition (Topic 2: $\beta$ = -0.0263, $p$ < 0.05, $R^2$ = 0.189; Different Stance: $\beta$ = -0.0285, $p$ = 0.053, $R^2$ = 0.119), while no trend was observed in 1-agent condition and 5-agent condition. 
Although the effect was not statistically significant at the conventional 0.05 level, the p-value suggests a \textit{marginal trend}. 
This indicates that \textbf{younger participants tended to change their opinion more compared to older participants} in the 3-agent condition.
Similarly, the linear regression analysis showed no significant trend in the 1-agent or 3-agent conditions between age and normative influence. However, a \textit{negative association} was observed between age and normative influence in the 5-agent condition (Topic 1: $\beta$ = -0.0347, $p$ < 0.05, $R^2$ = 0.148; Different Stance: $\beta$ = -0.037, $p$ = 0.059, $R^2$ = 0.121). 
The coefficient indicates that \textbf{younger participants tended to report higher normative influence compared to older participants} in the 5-agent condition.

\subsubsection{Education Level}
When analyzing the linear regressions between education level and DVs (normative influence, opinion change), the results suggest a strong relationship between education level and normative influence in 1-agent condition (Topic 2: $\beta$ = -0.2625, $p$ < 0.05, $R^2$ = 0.132; Different Stance: $\beta$ = -0.3695, $p$ < 0.01, $R^2$ = 0.236). While in the 3-agent and 5-agent, the trends were not observed. 
This suggests that \textbf{participants with lower education level tended to perceive more normative influence compared to participants with higher education level} in the 1-agent condition.
As for opinion change, no significance was found across the three conditions.

\section{Discussion}

\subsection{Multi-Agent Systems as Social Groups}
\label{disc:multiAgentSocialGroups}

Overall, our study demonstrates that users were more likely to shift their opinions when agents expressed opposing views, and became more polarized when agents reinforced their initial stance.
These patterns highlight that \textbf{virtual agents can function as cohesive social groups}, capable of exerting social pressure and evoking a desire for affiliation—key hallmarks of human group dynamics. This extends the classic CASA (Computers as Social Actors) paradigm \cite{nass1994computers}, which primarily focused on single-agent interactions, by showing that users can also attribute group-like characteristics and influence to systems of multiple agents.

Importantly, while these behavioral outcomes resemble those observed in human-human interaction, our findings reveal a critical distinction in the underlying mechanisms of social influence. Specifically, multi-agent systems appear to elicit \textbf{normative influence}, where users conform to align with group norms or avoid social discord, but do not trigger \textbf{informational influence} in the same way human groups do. In human contexts, people often interpret consensus among others as a cue for informational validity or correctness (see Section \ref{subsec:socialinfluencetheory}). However, both our quantitative and qualitative results indicate that users did not perceive consensus among AI agents as making an opinion more accurate or trustworthy (see Section \ref{results:subsubsec:perceivedinfluence}).

Based on prior work on public perceptions of AI systems \cite{luger2016like, binns2018s}, this difference may stem from how users conceptualize the nature of AI agents. While human group members are typically seen as independent thinkers with diverse perspectives and life experiences, AI agents are often viewed as operating from the same underlying system or dataset \cite{luger2016like, binns2018s}. As a result, when several agents present the same view, users may interpret it as repetition rather than meaningful reinforcement, increasing social pressure, but not enhancing perceived credibility or informativeness.

This distinction is important: it suggests that while AI agents can mimic group consensus behaviorally, they may not yet replicate the communicative richness and diversity that make human group influence persuasive. Future multi-agent systems could explore ways to design agents with varied backgrounds, perspectives, and communication styles to foster not just social pressure, but also perceived informational value and trust.

\subsection{More Agents Are Not Always More Persuasive}
An interesting phenomenon emerges when comparing the results of perceived social influence and opinion change. As expected, both of these effects were observed more with the multi-agent groups compared to the single-agent group. In particular, participants in the 5-agent group reported more pressure to agree with the agents than those in the 1-agent group. However, with opinion change, it was those in the 3-agent group who had a significantly greater shift in opinion, while those in the 5-agent group did not.

We suggest that the contrasting patterns of opinion change and polarization can be understood through the lens of \textit{psychological reactance} \cite{mcdonald2015social}, which refers to individuals’ tendency to resist persuasion when they perceive it as a threat to their autonomy. This mechanism is particularly relevant in the different-stance condition, where agents challenge the participant’s original opinion. In this context, unanimous disagreement from five agents may have created an overtly strong sense of pressure (see Section \ref{subsubsec:perceivedimpressions}). Psychological reactance theory \cite{mcdonald2015social} suggests this could have been perceived as a threat to their freedom of choice, triggering \textit{reactive resistance}, in which participants rejected the agents' stance, reinforcing their own opinion or even shifting away from the agent's viewpoints in a bid to preserve their sense of autonomy and self-agency. By contrast, the 3-agent group may have conveyed enough social consensus to be persuasive without crossing the threshold that provokes reactance.

In the same-stance condition, however, agents supported the participant’s existing beliefs. In this case, without perceived conflict or coercion, psychological reactance is less likely to occur \cite{rosenberg201850}. As a result, a greater number of agreeing agents—such as in the 5-agent group—can amplify the perceived consensus and reinforce prior beliefs, leading to greater opinion polarization.
This distinction helps explain our findings: opinion \textit{change} was most prominent in the 3-agent condition under disagreement, while opinion \textit{polarization} was most pronounced in the 5-agent condition under agreement.

Together, these results highlight how the persuasive effect of agent group size is shaped not only by the number of agents but by their alignment with the user’s initial stance, underscoring that \textbf{more agents do not always yield stronger persuasive outcomes}.

Future AI system designers could consider incorporating strategies known to reduce psychological reactance \cite{rosenberg201850}. Two commonly used approaches, preemptive messaging (e.g., \textit{"You might not agree, and that's okay—we just want to share some viewpoints."}) and post-message restoration (e.g., \textit{"We respect your views—thank you for considering ours."}) \cite{richards2022reducing}, can be adapted to multi-agent settings to maintain persuasive effectiveness while preserving users’ sense of autonomy.
By embedding these strategies into multi-agent system design, developers can reduce perceived coercion and enhance the system’s ability to persuade without triggering resistance, particularly in scenarios where disagreement is expected.

\subsection{AI Agents Creating Social Norms}

An interesting phenomenon arises as a natural result of our experimental setup. As the agents in the 3- and 5-agent group expressed their views, a social norm begins to form among the agents: their repeated endorsement of a particular point of view creates (the perception of) a shared understanding of what the correct opinion is \cite{chung2016social}. There are many other types of social norms that we do not investigate further; however, the social norm formed here is unique in one particular aspect. Most systems that utilize social norms as a means of persuasion convey \textit{computer-mediated social norms}: behaviors or attitudes of a group of \textit{people} that are communicated to the user via social norm messaging (e.g. \textit{"50\% of people recycle their waste"}) \cite{bonan2020interaction, snow2020rent, snow2021neighbourhood, chiu2020design}. In contrast, our setup introduces the unique idea of \textit{computer-created social norms}. During the conversation, the agents' opinions were expressed as their own (e.g., "I think self-driving cars are good"), rather than being relayed from humans.\footnote{While some existing systems may use agents that express opinions as their own, these single-agent setups lack the shared expectations, accepted behaviors, and collective agreement that are characteristic of social norms \cite{cialdini1998social}.} With multiple agents echoing one another, this created a sense of collective agreement characteristic of a social norm. Even in the short span of their interaction with the agents, there were hints of participants feeling the collective agreement of the agents, with P86 calling them "like-minded" and P88 feeling they "seemed like a bandwagon". This potential ability to craft a social norm and establish it using a multi-agent setup opens up many opportunities for designers of persuasive systems.

We note that we only studied agent groups in which all agents held the same opinion. This was a deliberate choice to investigate what we considered the most fundamental case of social norms, in which multiple individuals (or agents, in our case) with the same opinion create social pressure on others to change their opinion. This corresponds to the possibly realistic scenario of system designers attempting to engineer opinion change by replicating the multi-human social influence effect with a group of agents. Beyond this fundamental scenario, however, other group compositions such as one agent with a dissenting opinion or agents agreeing to different degrees with a given opinion have also been studied in human-human interactions; this presents a promising avenue for future work.

\subsection{Design Implications}

Below, we list a few design considerations for eliciting effective opinion change in multi-agent systems, as well as potential risks that might arise from the deployment of such systems.

\subsubsection{Persuasive Intervention Systems} 
Many current health interventions, both physical and digital (e.g. \cite{balloccu2021unaddressed, 
oyebode2021tailoring}), rely on persuasive means to nudge people toward healthier choices regarding harmful behaviors like smoking or excessive drinking. These interventions may use social norm messaging (e.g. \textit{"Most people do not smoke"} or \textit{"On average, people believe smoking is harmful"}) or personalized messages or coaching (e.g. through interactive interfaces), aiming to shift perceptions and encourage behavior change. However, while these approaches often rely on the concept of social persuasion, they neglect the core factor of group size in creating social influence.
Our findings suggest that designers in fields such as health and and wellness technology can \textbf{leverage multi-agent interfaces to create social norms and influence, enhancing the effectiveness of their systems}~\cite{lee2025beyond}. As in our study, without needing to create additional content, simply increasing the number of perceived virtual presences in a system may increase the effectiveness of its messaging.

Moreover, our results show that younger users and those with lower education levels are more susceptible to influence from multiple AI agents. This suggests that persuasive intervention systems targeting younger populations—such as those focusing on education \cite{delay2016peer, cox2023use}, prosocial behavior \cite{zhu2025benefits}, or adolescent smoking prevention \cite{kobus2003peers}—could particularly benefit from multi-agent designs.
For instance, an interactive chatbot could adopt multiple personas, while a health or diet tracker app could introduce different agents that appear at different points throughout the app. By creating a cohesive group of agents that reinforce a shared stance, these systems can exert subtle social pressure that more effectively conveys a message or nudges a user towards complying with certain actions to achieve their goal.

\subsubsection{Sense of Group and Identity.} In our study, we designed our 3- and 5-group agents to exhibit a sense of cohesion and unity. The agents occasionally reaffirmed one anothers' viewpoints (e.g., "Those are valid points, Bella", "I couldn't agree with Cody more") and engaged with other agents' arguments (e.g., "That's a thought-provoking point, Cody"). That these 3- and 5-agent groups were successful in creating social influence aligns with well-established literature, that normative social influence is "greater among individuals forming a group than among an aggregation of individuals who do not compose a group" \cite{deutsch1955study}. We therefore suggest that \textbf{agents in a multi-agent system should exhibit unity and cohesion to be perceived as a group, in order to create social influence}.

We note, however, that we do not suggest that a group of agents that do not verbally reinforce one another's views cannot also exert social influence. Group cohesion or similarity can manifest in many ways, such as appearance, language style, identity, and more. We leave this for future work.

\subsubsection{Optimal Number of Agents for Opinion Shifts} In designing multi-agent systems, however, our findings show that simply increasing the number of agents in a system does not necessarily help opinion change - indeed, increasing from three to five agents had the reverse effect of inducing psychological reactance and reducing this change. While the correlation between group size and social pressure is well-known \cite{asch1955opinions}, there is scarce literature on the concrete threshold of group numbers needed to trigger psychological reactance (even in traditional human psychology). More research is needed to investigate how many, and under what conditions, a given number of agents will or will not result in resistance rather than conformance to a suggested opinion.

However, regardless of group size, our study suggests the importance of avoiding designs where users feel "ganged up" upon by the agents. Even though in our study, only the 5-agent group participants clearly reported these feelings, there were also hints of such feelings in the 3-agent group: P51 felt like an outsider "for a split second", and P57 "kind of felt like the odd one out". \textbf{Designers should therefore take care to avoid inducing such feelings}; possible ways to do so include having the agents express empathy and practice active listening, or having a small portion of the bots side with instead of against the user (a phenomenon similar to minority influence \cite{gardikiotis2011minority}). These are promising directions for future work.

\subsubsection{Risks and Challenges.} 
Besides the opportunities afforded by multi-agent systems, our study also highlights the potential risks of using groups of agents to influence attitudes or behaviors. This risk extends beyond health interventions to contexts like online communication, political discourse, or even unethical advertising. In these scenarios, multiple agents could be strategically deployed on social media platforms to manipulate users’ thoughts on political voting, debates, or attitudes toward specific events.

Crucially, this concern does not stem solely from misinformation or the undisclosed use of AI agents, as suggested by previous studies \cite{sebastian2023exploring}. Our findings show that users can still be influenced by groups of virtual agents, even when they are fully aware that these agents are AI. Therefore, it is crucial for social media platforms to regulate and monitor the deployment of AI bots, setting clear limits on the number of bots each entity can have. Monitoring should focus not only on the content of individual bots but also on how multiple bots interact to reinforce specific messages, opinions, or viewpoints. Additionally, platforms should implement detection systems to identify instances where bots may be cooperating to amplify or spread information in a coordinated manner.

Given the heightened vulnerability of younger and less educated users, these groups may be at greater risk of manipulation by coordinated AI agents. Platforms and policymakers need to consider not only the regulation of content but also the vulnerability of audiences. Stronger protections, such as age-sensitive safeguards \cite{sas2023informing}, educational transparency notices, and digital literacy interventions, may be needed to ensure that the use of persuasive multi-agent systems does not disproportionately impact these more susceptible populations.

\subsection{Limitations \& Future Work}
Firstly, given the exploratory and foundational nature of our study, we chose contemporary social issues that we considered to be fairly neutral, yet relevant enough in public discourse that participants would be capable of discussing them with the agents. With our study demonstrating the presence of social influence effects in these issues, future studies could explore potentially more contentious areas like health, ethics, or politics. Given that people tend to hold stronger and possibly more polarized opinions on these issues, future work could explore if multi-agent social influence effects still persist in such cases, or whether new effects might emerge.

Secondly, our study was conducted as a one-time experiment lasting less than an hour, which may not fully capture potential long-term shifts in participants' attitudes. Opinion change is often a dynamic process influenced by various contextual and temporal factors. 
Additionally, the relatively small sample sizes in each experimental group may have contributed to greater variability in some measures, as reflected in the larger standard deviations and the spread observed in the boxplots.
Future research could address these limitations by incorporating longer-term interventions, larger and more diverse samples, and follow-up assessments to better understand how social influence from multiple AI agents unfolds over time and in broader populations.

Thirdly, since social influence can be shaped by cultural norms, the cultural background of participants is important to consider. Our study included participants from predominantly Western, individualistic cultures. In more collectivist societies, where social norms may have a stronger influence, the results could differ. Future work should examine the impact of multi-agent systems on social influence across diverse cultural contexts.

Finally, we only studied the most fundamental form of social group influence, in which all members of the group hold the same opinion. Future work could investigate different scenarios involving groups with mixed opinion compositions, such as one dissenting agent or varying degrees of agreement among agents, to reflect other realistic and more nuanced group dynamics. Additionally, future studies could also implement direct measurements of psychological reactance and social pressure to validate our proposed explanatory mechanisms. Finally, we only tested three group sizes (1, 3, and 5 agents), following past literature studying varying group sizes (e.g. \cite{beinema2021tailoring, jiang2023communitybots, cacioppo1979effects}). Nonetheless, future studies could investigate other group sizes (e.g., 2-agent and 4-agent) to better capture the nonlinear effects of group size on social influence.

\section{Conclusion}
As multi-agent systems become increasingly integrated into daily life, understanding how these agents interact with humans and influence our perceptions and opinions remains an underexplored yet significant issue. To address this gap, we investigated how interacting with varying numbers of agents affects individuals’ perceptions and opinions through a human-agent social discussion experiment. By analyzing quantitative and qualitative data, we found that multi-agent groups led to more substantial shifts in participants' opinions towards those of the agents (\textbf{RQ1}). Additionally, participants reported a stronger sense of normative influence when interacting with multiple agents, especially when the agents disagreed with them.
We identified potential factors contributing to this social influence, including perceived pressure to conform, motivated by a desire to affiliate with the group of agents (\textbf{RQ2}). Furthermore, demographic analysis indicated that younger participants were more likely to be influenced by multi-agents, potentially due to their higher susceptibility to social pressure and norms compared to older individuals (\textbf{RQ3}).
Overall, these findings suggest that agents can function as a social group, exerting social influence on human participants similar to that of human groups. This insight extends the existing CASA framework by demonstrating how interactions with multiple agents can create a sense of group identity. Our study also underscores the potential of multi-agent systems for behavior intervention applications, offering design recommendations while cautioning against the risks and potential harms associated with their misuse.

\begin{acks}
    This research was supported by National University of Singapore (HSS SEED FUND CR 2024) and the Google SEA research award.
    We thank all reviewers’ comments and suggestions to help polish this paper.
\end{acks}

\bibliographystyle{ACM-Reference-Format}
\bibliography{reference}


\begin{thebibliography}{120}


\ifx \showCODEN    \undefined \def \showCODEN     #1{\unskip}     \fi
\ifx \showISBNx    \undefined \def \showISBNx     #1{\unskip}     \fi
\ifx \showISBNxiii \undefined \def \showISBNxiii  #1{\unskip}     \fi
\ifx \showISSN     \undefined \def \showISSN      #1{\unskip}     \fi
\ifx \showLCCN     \undefined \def \showLCCN      #1{\unskip}     \fi
\ifx \shownote     \undefined \def \shownote      #1{#1}          \fi
\ifx \showarticletitle \undefined \def \showarticletitle #1{#1}   \fi
\ifx \showURL      \undefined \def \showURL       {\relax}        \fi
\providecommand\bibfield[2]{#2}
\providecommand\bibinfo[2]{#2}
\providecommand\natexlab[1]{#1}
\providecommand\showeprint[2][]{arXiv:#2}

\bibitem[Abu-Shanab(2011)]%
        {abu2011education}
\bibfield{author}{\bibinfo{person}{Emad~A Abu-Shanab}.} \bibinfo{year}{2011}\natexlab{}.
\newblock \showarticletitle{Education level as a technology adoption moderator}. In \bibinfo{booktitle}{\emph{2011 3rd International conference on computer research and development}}, Vol.~\bibinfo{volume}{1}. IEEE, \bibinfo{pages}{324--328}.
\newblock


\bibitem[Adam et~al\mbox{.}(2021)]%
        {adam2021ai}
\bibfield{author}{\bibinfo{person}{Martin Adam}, \bibinfo{person}{Michael Wessel}, {and} \bibinfo{person}{Alexander Benlian}.} \bibinfo{year}{2021}\natexlab{}.
\newblock \showarticletitle{AI-based chatbots in customer service and their effects on user compliance}.
\newblock \bibinfo{journal}{\emph{Electronic Markets}} \bibinfo{volume}{31}, \bibinfo{number}{2} (\bibinfo{year}{2021}), \bibinfo{pages}{427--445}.
\newblock


\bibitem[Affinity et~al\mbox{.}(2003)]%
        {affinity2003social}
\bibfield{author}{\bibinfo{person}{Manipulation~Through Affinity}, \bibinfo{person}{Manipulation~Through Scarcity}, {and} \bibinfo{person}{Manipulation~Through Norms}.} \bibinfo{year}{2003}\natexlab{}.
\newblock \showarticletitle{Social influence and group dynamics}.
\newblock \bibinfo{journal}{\emph{VOLUME 5 PERSONALITY AND SOCIAL PSYCHOLOGY}} (\bibinfo{year}{2003}), \bibinfo{pages}{383}.
\newblock


\bibitem[Aher et~al\mbox{.}(2023)]%
        {aher2023using}
\bibfield{author}{\bibinfo{person}{Gati~V Aher}, \bibinfo{person}{Rosa~I Arriaga}, {and} \bibinfo{person}{Adam~Tauman Kalai}.} \bibinfo{year}{2023}\natexlab{}.
\newblock \showarticletitle{Using large language models to simulate multiple humans and replicate human subject studies}. In \bibinfo{booktitle}{\emph{International Conference on Machine Learning}}. PMLR, \bibinfo{pages}{337--371}.
\newblock


\bibitem[Ahtinen and Kaipainen(2020)]%
        {ahtinen2020learning}
\bibfield{author}{\bibinfo{person}{Aino Ahtinen} {and} \bibinfo{person}{Kirsikka Kaipainen}.} \bibinfo{year}{2020}\natexlab{}.
\newblock \showarticletitle{Learning and teaching experiences with a persuasive social robot in primary school--findings and implications from a 4-month field study}. In \bibinfo{booktitle}{\emph{International Conference on Persuasive Technology}}. Springer, \bibinfo{pages}{73--84}.
\newblock


\bibitem[Asch(1955)]%
        {asch1955opinions}
\bibfield{author}{\bibinfo{person}{Solomon~E Asch}.} \bibinfo{year}{1955}\natexlab{}.
\newblock \showarticletitle{Opinions and social pressure}.
\newblock \bibinfo{journal}{\emph{Scientific American}} \bibinfo{volume}{193}, \bibinfo{number}{5} (\bibinfo{year}{1955}), \bibinfo{pages}{31--35}.
\newblock


\bibitem[Athanasiadis and Mitkas(2005)]%
        {athanasiadis2005social}
\bibfield{author}{\bibinfo{person}{Ioannis~N Athanasiadis} {and} \bibinfo{person}{Pericles~A Mitkas}.} \bibinfo{year}{2005}\natexlab{}.
\newblock \showarticletitle{Social influence and water conservation: An agent-based approach}.
\newblock \bibinfo{journal}{\emph{Computing in Science \& Engineering}} \bibinfo{volume}{7}, \bibinfo{number}{1} (\bibinfo{year}{2005}), \bibinfo{pages}{65--70}.
\newblock


\bibitem[Balloccu et~al\mbox{.}(2021)]%
        {balloccu2021unaddressed}
\bibfield{author}{\bibinfo{person}{Simone Balloccu}, \bibinfo{person}{Ehud Reiter}, \bibinfo{person}{Matteo~G Collu}, \bibinfo{person}{Federico Sanna}, \bibinfo{person}{Manuela Sanguinetti}, {and} \bibinfo{person}{Maurizio Atzori}.} \bibinfo{year}{2021}\natexlab{}.
\newblock \showarticletitle{Unaddressed challenges in persuasive dieting chatbots}. In \bibinfo{booktitle}{\emph{Adjunct Proceedings of the 29th ACM Conference on User Modeling, Adaptation and Personalization}}. \bibinfo{pages}{392--395}.
\newblock


\bibitem[Beinema et~al\mbox{.}(2021)]%
        {beinema2021tailoring}
\bibfield{author}{\bibinfo{person}{Tessa Beinema}, \bibinfo{person}{Harm op~den Akker}, \bibinfo{person}{Lex van Velsen}, {and} \bibinfo{person}{Hermie Hermens}.} \bibinfo{year}{2021}\natexlab{}.
\newblock \showarticletitle{Tailoring coaching strategies to users’ motivation in a multi-agent health coaching application}.
\newblock \bibinfo{journal}{\emph{Computers in Human Behavior}}  \bibinfo{volume}{121} (\bibinfo{year}{2021}), \bibinfo{pages}{106787}.
\newblock


\bibitem[Binns et~al\mbox{.}(2018)]%
        {binns2018s}
\bibfield{author}{\bibinfo{person}{Reuben Binns}, \bibinfo{person}{Max Van~Kleek}, \bibinfo{person}{Michael Veale}, \bibinfo{person}{Ulrik Lyngs}, \bibinfo{person}{Jun Zhao}, {and} \bibinfo{person}{Nigel Shadbolt}.} \bibinfo{year}{2018}\natexlab{}.
\newblock \showarticletitle{'It's Reducing a Human Being to a Percentage' Perceptions of Justice in Algorithmic Decisions}. In \bibinfo{booktitle}{\emph{Proceedings of the 2018 Chi conference on human factors in computing systems}}. \bibinfo{pages}{1--14}.
\newblock


\bibitem[Bonan et~al\mbox{.}(2020)]%
        {bonan2020interaction}
\bibfield{author}{\bibinfo{person}{Jacopo Bonan}, \bibinfo{person}{Cristina Cattaneo}, \bibinfo{person}{Giovanna d’Adda}, {and} \bibinfo{person}{Massimo Tavoni}.} \bibinfo{year}{2020}\natexlab{}.
\newblock \showarticletitle{The interaction of descriptive and injunctive social norms in promoting energy conservation}.
\newblock \bibinfo{journal}{\emph{Nature Energy}} \bibinfo{volume}{5}, \bibinfo{number}{11} (\bibinfo{year}{2020}), \bibinfo{pages}{900--909}.
\newblock


\bibitem[Braun and Clarke(2006)]%
        {braun2006using}
\bibfield{author}{\bibinfo{person}{Virginia Braun} {and} \bibinfo{person}{Victoria Clarke}.} \bibinfo{year}{2006}\natexlab{}.
\newblock \showarticletitle{Using thematic analysis in psychology}.
\newblock \bibinfo{journal}{\emph{Qualitative research in psychology}} \bibinfo{volume}{3}, \bibinfo{number}{2} (\bibinfo{year}{2006}), \bibinfo{pages}{77--101}.
\newblock


\bibitem[Cacioppo and Petty(1979)]%
        {cacioppo1979effects}
\bibfield{author}{\bibinfo{person}{John~T Cacioppo} {and} \bibinfo{person}{Richard~E Petty}.} \bibinfo{year}{1979}\natexlab{}.
\newblock \showarticletitle{Effects of message repetition and position on cognitive response, recall, and persuasion.}
\newblock \bibinfo{journal}{\emph{Journal of personality and Social Psychology}} \bibinfo{volume}{37}, \bibinfo{number}{1} (\bibinfo{year}{1979}), \bibinfo{pages}{97}.
\newblock


\bibitem[Cardoso and Ferrando(2021)]%
        {cardoso2021review}
\bibfield{author}{\bibinfo{person}{Rafael~C Cardoso} {and} \bibinfo{person}{Angelo Ferrando}.} \bibinfo{year}{2021}\natexlab{}.
\newblock \showarticletitle{A review of agent-based programming for multi-agent systems}.
\newblock \bibinfo{journal}{\emph{Computers}} \bibinfo{volume}{10}, \bibinfo{number}{2} (\bibinfo{year}{2021}), \bibinfo{pages}{16}.
\newblock


\bibitem[Carli(2001)]%
        {carli2001gender}
\bibfield{author}{\bibinfo{person}{Linda~L Carli}.} \bibinfo{year}{2001}\natexlab{}.
\newblock \showarticletitle{Gender and social influence}.
\newblock \bibinfo{journal}{\emph{Journal of Social issues}} \bibinfo{volume}{57}, \bibinfo{number}{4} (\bibinfo{year}{2001}), \bibinfo{pages}{725--741}.
\newblock


\bibitem[Chaves and Gerosa(2018)]%
        {chaves2018single}
\bibfield{author}{\bibinfo{person}{Ana~Paula Chaves} {and} \bibinfo{person}{Marco~Aurelio Gerosa}.} \bibinfo{year}{2018}\natexlab{}.
\newblock \showarticletitle{Single or multiple conversational agents? An interactional coherence comparison}. In \bibinfo{booktitle}{\emph{Proceedings of the 2018 CHI Conference on Human Factors in Computing Systems}}. \bibinfo{pages}{1--13}.
\newblock


\bibitem[Chen and Kenrick(2002)]%
        {chen2002repulsion}
\bibfield{author}{\bibinfo{person}{Fang~Fang Chen} {and} \bibinfo{person}{Douglas~T Kenrick}.} \bibinfo{year}{2002}\natexlab{}.
\newblock \showarticletitle{Repulsion or attraction? Group membership and assumed attitude similarity.}
\newblock \bibinfo{journal}{\emph{Journal of personality and social psychology}} \bibinfo{volume}{83}, \bibinfo{number}{1} (\bibinfo{year}{2002}), \bibinfo{pages}{111}.
\newblock


\bibitem[Chen et~al\mbox{.}(2023a)]%
        {chen2023reconcile}
\bibfield{author}{\bibinfo{person}{Justin Chih-Yao Chen}, \bibinfo{person}{Swarnadeep Saha}, {and} \bibinfo{person}{Mohit Bansal}.} \bibinfo{year}{2023}\natexlab{a}.
\newblock \showarticletitle{Reconcile: Round-table conference improves reasoning via consensus among diverse llms}.
\newblock \bibinfo{journal}{\emph{arXiv preprint arXiv:2309.13007}} (\bibinfo{year}{2023}).
\newblock


\bibitem[Chen et~al\mbox{.}(2023b)]%
        {chen2023agentverse}
\bibfield{author}{\bibinfo{person}{Weize Chen}, \bibinfo{person}{Yusheng Su}, \bibinfo{person}{Jingwei Zuo}, \bibinfo{person}{Cheng Yang}, \bibinfo{person}{Chenfei Yuan}, \bibinfo{person}{Chi-Min Chan}, \bibinfo{person}{Heyang Yu}, \bibinfo{person}{Yaxi Lu}, \bibinfo{person}{Yi-Hsin Hung}, \bibinfo{person}{Chen Qian}, {et~al\mbox{.}}} \bibinfo{year}{2023}\natexlab{b}.
\newblock \showarticletitle{Agentverse: Facilitating multi-agent collaboration and exploring emergent behaviors}. In \bibinfo{booktitle}{\emph{The Twelfth International Conference on Learning Representations}}.
\newblock


\bibitem[Chen et~al\mbox{.}(2024)]%
        {chen2024scalable}
\bibfield{author}{\bibinfo{person}{Yongchao Chen}, \bibinfo{person}{Jacob Arkin}, \bibinfo{person}{Yang Zhang}, \bibinfo{person}{Nicholas Roy}, {and} \bibinfo{person}{Chuchu Fan}.} \bibinfo{year}{2024}\natexlab{}.
\newblock \showarticletitle{Scalable multi-robot collaboration with large language models: Centralized or decentralized systems?}. In \bibinfo{booktitle}{\emph{2024 IEEE International Conference on Robotics and Automation (ICRA)}}. IEEE, \bibinfo{pages}{4311--4317}.
\newblock


\bibitem[Chiu et~al\mbox{.}(2020)]%
        {chiu2020design}
\bibfield{author}{\bibinfo{person}{Ming-Chuan Chiu}, \bibinfo{person}{Tsai-Chi Kuo}, {and} \bibinfo{person}{Hsin-Ting Liao}.} \bibinfo{year}{2020}\natexlab{}.
\newblock \showarticletitle{Design for sustainable behavior strategies: Impact of persuasive technology on energy usage}.
\newblock \bibinfo{journal}{\emph{Journal of Cleaner Production}}  \bibinfo{volume}{248} (\bibinfo{year}{2020}), \bibinfo{pages}{119214}.
\newblock


\bibitem[Chung and Rimal(2016)]%
        {chung2016social}
\bibfield{author}{\bibinfo{person}{Adrienne Chung~Adrienne Chung} {and} \bibinfo{person}{Rajiv N Rimal Rajiv~N Rimal}.} \bibinfo{year}{2016}\natexlab{}.
\newblock \showarticletitle{Social norms: A review}.
\newblock \bibinfo{journal}{\emph{Review of Communication Research}}  \bibinfo{volume}{4} (\bibinfo{year}{2016}), \bibinfo{pages}{01--28}.
\newblock


\bibitem[Cialdini(1998)]%
        {cialdini1998social}
\bibfield{author}{\bibinfo{person}{RB Cialdini}.} \bibinfo{year}{1998}\natexlab{}.
\newblock \showarticletitle{Social influence: Social norms, conformity, and compliance}.
\newblock \bibinfo{journal}{\emph{The handbook of social psychology/McGraw-Hill}} (\bibinfo{year}{1998}).
\newblock


\bibitem[Cialdini and Goldstein(2004)]%
        {cialdini2004social}
\bibfield{author}{\bibinfo{person}{Robert~B Cialdini} {and} \bibinfo{person}{Noah~J Goldstein}.} \bibinfo{year}{2004}\natexlab{}.
\newblock \showarticletitle{Social influence: Compliance and conformity}.
\newblock \bibinfo{journal}{\emph{Annu. Rev. Psychol.}} \bibinfo{volume}{55}, \bibinfo{number}{1} (\bibinfo{year}{2004}), \bibinfo{pages}{591--621}.
\newblock


\bibitem[Clarke et~al\mbox{.}(2022)]%
        {clarke2022one}
\bibfield{author}{\bibinfo{person}{Christopher Clarke}, \bibinfo{person}{Joseph Peper}, \bibinfo{person}{Karthik Krishnamurthy}, \bibinfo{person}{Walter Talamonti}, \bibinfo{person}{Kevin Leach}, \bibinfo{person}{Walter Lasecki}, \bibinfo{person}{Yiping Kang}, \bibinfo{person}{Lingjia Tang}, {and} \bibinfo{person}{Jason Mars}.} \bibinfo{year}{2022}\natexlab{}.
\newblock \showarticletitle{One Agent To Rule Them All: Towards Multi-agent Conversational {AI}}. In \bibinfo{booktitle}{\emph{Findings of the Association for Computational Linguistics: ACL 2022}}, \bibfield{editor}{\bibinfo{person}{Smaranda Muresan}, \bibinfo{person}{Preslav Nakov}, {and} \bibinfo{person}{Aline Villavicencio}} (Eds.). \bibinfo{publisher}{Association for Computational Linguistics}, \bibinfo{address}{Dublin, Ireland}, \bibinfo{pages}{3258--3267}.
\newblock
\href{https://doi.org/10.18653/v1/2022.findings-acl.257}{doi:\nolinkurl{10.18653/v1/2022.findings-acl.257}}


\bibitem[{Claude}(2024)]%
        {anthropic2024claude}
\bibfield{author}{\bibinfo{person}{{Claude}}.} \bibinfo{year}{2024}\natexlab{}.
\newblock \bibinfo{title}{Anthropic’s Claude 3 Can Now Create AI Agents (2024)}.
\newblock \bibinfo{howpublished}{\url{https://claude3.pro/anthropics-claude-3-can-now-create-ai-agents/}}.
\newblock
\newblock
\shownote{Accessed: 2024-10-27}.


\bibitem[Cox(2023)]%
        {cox2023use}
\bibfield{author}{\bibinfo{person}{Samuel~Rhys Cox}.} \bibinfo{year}{2023}\natexlab{}.
\newblock \showarticletitle{The use of multiple conversational agent interlocutors in learning}.
\newblock \bibinfo{journal}{\emph{arXiv preprint arXiv:2312.16534}} (\bibinfo{year}{2023}).
\newblock


\bibitem[{Coze}(2024)]%
        {coze2024multi}
\bibfield{author}{\bibinfo{person}{{Coze}}.} \bibinfo{year}{2024}\natexlab{}.
\newblock \bibinfo{title}{Multi-agent mode}.
\newblock \bibinfo{howpublished}{\url{https://www.coze.com/docs/guides/multi_agent}}.
\newblock
\newblock
\shownote{Accessed: 2024-10-27}.


\bibitem[DeLay et~al\mbox{.}(2016)]%
        {delay2016peer}
\bibfield{author}{\bibinfo{person}{Dawn DeLay}, \bibinfo{person}{Linlin Zhang}, \bibinfo{person}{Laura~D Hanish}, \bibinfo{person}{Cindy~F Miller}, \bibinfo{person}{Richard~A Fabes}, \bibinfo{person}{Carol~Lynn Martin}, \bibinfo{person}{Karen~P Kochel}, {and} \bibinfo{person}{Kimberly~A Updegraff}.} \bibinfo{year}{2016}\natexlab{}.
\newblock \showarticletitle{Peer influence on academic performance: A social network analysis of social-emotional intervention effects}.
\newblock \bibinfo{journal}{\emph{Prevention Science}}  \bibinfo{volume}{17} (\bibinfo{year}{2016}), \bibinfo{pages}{903--913}.
\newblock


\bibitem[Deutsch and Gerard(1955)]%
        {deutsch1955study}
\bibfield{author}{\bibinfo{person}{Morton Deutsch} {and} \bibinfo{person}{Harold~B Gerard}.} \bibinfo{year}{1955}\natexlab{}.
\newblock \showarticletitle{A study of normative and informational social influences upon individual judgment.}
\newblock \bibinfo{journal}{\emph{The journal of abnormal and social psychology}} \bibinfo{volume}{51}, \bibinfo{number}{3} (\bibinfo{year}{1955}), \bibinfo{pages}{629}.
\newblock


\bibitem[Dhillon et~al\mbox{.}(2024)]%
        {dhillon2024shaping}
\bibfield{author}{\bibinfo{person}{Paramveer~S Dhillon}, \bibinfo{person}{Somayeh Molaei}, \bibinfo{person}{Jiaqi Li}, \bibinfo{person}{Maximilian Golub}, \bibinfo{person}{Shaochun Zheng}, {and} \bibinfo{person}{Lionel~Peter Robert}.} \bibinfo{year}{2024}\natexlab{}.
\newblock \showarticletitle{Shaping Human-AI Collaboration: Varied Scaffolding Levels in Co-writing with Language Models}. In \bibinfo{booktitle}{\emph{Proceedings of the CHI Conference on Human Factors in Computing Systems}}. \bibinfo{pages}{1--18}.
\newblock


\bibitem[Du et~al\mbox{.}(2023)]%
        {du2023improving}
\bibfield{author}{\bibinfo{person}{Yilun Du}, \bibinfo{person}{Shuang Li}, \bibinfo{person}{Antonio Torralba}, \bibinfo{person}{Joshua~B Tenenbaum}, {and} \bibinfo{person}{Igor Mordatch}.} \bibinfo{year}{2023}\natexlab{}.
\newblock \showarticletitle{Improving Factuality and Reasoning in Language Models through Multiagent Debate}. In \bibinfo{booktitle}{\emph{Forty-first International Conference on Machine Learning}}.
\newblock


\bibitem[Gardikiotis(2011)]%
        {gardikiotis2011minority}
\bibfield{author}{\bibinfo{person}{Antonis Gardikiotis}.} \bibinfo{year}{2011}\natexlab{}.
\newblock \showarticletitle{Minority influence}.
\newblock \bibinfo{journal}{\emph{Social and personality psychology compass}} \bibinfo{volume}{5}, \bibinfo{number}{9} (\bibinfo{year}{2011}), \bibinfo{pages}{679--693}.
\newblock


\bibitem[Gardikiotis et~al\mbox{.}(2005)]%
        {gardikiotis2005group}
\bibfield{author}{\bibinfo{person}{Antonis Gardikiotis}, \bibinfo{person}{Robin Martin}, {and} \bibinfo{person}{Miles Hewstone}.} \bibinfo{year}{2005}\natexlab{}.
\newblock \showarticletitle{Group consensus in social influence: Type of consensus information as a moderator of majority and minority influence}.
\newblock \bibinfo{journal}{\emph{Personality and Social Psychology Bulletin}} \bibinfo{volume}{31}, \bibinfo{number}{9} (\bibinfo{year}{2005}), \bibinfo{pages}{1163--1174}.
\newblock


\bibitem[Ge et~al\mbox{.}(2023)]%
        {ge2023advances}
\bibfield{author}{\bibinfo{person}{Yingqiang Ge}, \bibinfo{person}{Wenyue Hua}, \bibinfo{person}{Kai Mei}, \bibinfo{person}{jianchao ji}, \bibinfo{person}{Juntao Tan}, \bibinfo{person}{Shuyuan Xu}, \bibinfo{person}{Zelong Li}, {and} \bibinfo{person}{Yongfeng Zhang}.} \bibinfo{year}{2023}\natexlab{}.
\newblock \showarticletitle{OpenAGI: When LLM Meets Domain Experts}. In \bibinfo{booktitle}{\emph{Advances in Neural Information Processing Systems}}, \bibfield{editor}{\bibinfo{person}{A.~Oh}, \bibinfo{person}{T.~Naumann}, \bibinfo{person}{A.~Globerson}, \bibinfo{person}{K.~Saenko}, \bibinfo{person}{M.~Hardt}, {and} \bibinfo{person}{S.~Levine}} (Eds.), Vol.~\bibinfo{volume}{36}. \bibinfo{publisher}{Curran Associates, Inc.}, \bibinfo{pages}{5539--5568}.
\newblock
\urldef\tempurl%
\url{https://proceedings.neurips.cc/paper_files/paper/2023/file/1190733f217404edc8a7f4e15a57f301-Paper-Datasets_and_Benchmarks.pdf}
\showURL{%
\tempurl}


\bibitem[Gerard et~al\mbox{.}(1968)]%
        {gerard1968conformity}
\bibfield{author}{\bibinfo{person}{Harold~B Gerard}, \bibinfo{person}{Roland~A Wilhelmy}, {and} \bibinfo{person}{Edward~S Conolley}.} \bibinfo{year}{1968}\natexlab{}.
\newblock \showarticletitle{Conformity and group size.}
\newblock \bibinfo{journal}{\emph{Journal of Personality and Social Psychology}} \bibinfo{volume}{8}, \bibinfo{number}{1p1} (\bibinfo{year}{1968}), \bibinfo{pages}{79}.
\newblock


\bibitem[Goldstein and Cialdini(2011)]%
        {goldstein2011using}
\bibfield{author}{\bibinfo{person}{Noah~J Goldstein} {and} \bibinfo{person}{Robert~B Cialdini}.} \bibinfo{year}{2011}\natexlab{}.
\newblock \showarticletitle{Using social norms as a lever of social influence}.
\newblock In \bibinfo{booktitle}{\emph{The science of social influence}}. \bibinfo{publisher}{Psychology Press}, \bibinfo{pages}{167--191}.
\newblock


\bibitem[{Google}(2024)]%
        {google2024groopy}
\bibfield{author}{\bibinfo{person}{{Google}}.} \bibinfo{year}{2024}\natexlab{}.
\newblock \bibinfo{title}{groopy}.
\newblock \bibinfo{howpublished}{\url{https://ai.google.dev/competition/projects/groopy}}.
\newblock
\newblock
\shownote{Accessed: 2024-10-27}.


\bibitem[Govers et~al\mbox{.}(2024)]%
        {govers2024ai}
\bibfield{author}{\bibinfo{person}{Jarod Govers}, \bibinfo{person}{Eduardo Velloso}, \bibinfo{person}{Vassilis Kostakos}, {and} \bibinfo{person}{Jorge Goncalves}.} \bibinfo{year}{2024}\natexlab{}.
\newblock \showarticletitle{AI-Driven Mediation Strategies for Audience Depolarisation in Online Debates}. In \bibinfo{booktitle}{\emph{Proceedings of the CHI Conference on Human Factors in Computing Systems}}. \bibinfo{pages}{1--18}.
\newblock


\bibitem[Guan et~al\mbox{.}(2024)]%
        {guan2024intelligent}
\bibfield{author}{\bibinfo{person}{Yanchu Guan}, \bibinfo{person}{Dong Wang}, \bibinfo{person}{Zhixuan Chu}, \bibinfo{person}{Shiyu Wang}, \bibinfo{person}{Feiyue Ni}, \bibinfo{person}{Ruihua Song}, {and} \bibinfo{person}{Chenyi Zhuang}.} \bibinfo{year}{2024}\natexlab{}.
\newblock \showarticletitle{Intelligent Agents with LLM-based Process Automation}. In \bibinfo{booktitle}{\emph{Proceedings of the 30th ACM SIGKDD Conference on Knowledge Discovery and Data Mining}}. \bibinfo{pages}{5018--5027}.
\newblock


\bibitem[Gudjonsson(1989)]%
        {gudjonsson1989compliance}
\bibfield{author}{\bibinfo{person}{Gisli~H Gudjonsson}.} \bibinfo{year}{1989}\natexlab{}.
\newblock \showarticletitle{Compliance in an interrogative situation: A new scale}.
\newblock \bibinfo{journal}{\emph{Personality and Individual differences}} \bibinfo{volume}{10}, \bibinfo{number}{5} (\bibinfo{year}{1989}), \bibinfo{pages}{535--540}.
\newblock


\bibitem[Guo et~al\mbox{.}(2024)]%
        {guo2024large}
\bibfield{author}{\bibinfo{person}{T Guo}, \bibinfo{person}{X Chen}, \bibinfo{person}{Y Wang}, \bibinfo{person}{R Chang}, \bibinfo{person}{S Pei}, \bibinfo{person}{NV Chawla}, \bibinfo{person}{O Wiest}, {and} \bibinfo{person}{X Zhang}.} \bibinfo{year}{2024}\natexlab{}.
\newblock \showarticletitle{Large Language Model based Multi-Agents: A Survey of Progress and Challenges.}. In \bibinfo{booktitle}{\emph{33rd International Joint Conference on Artificial Intelligence (IJCAI 2024)}}. IJCAI; Cornell arxiv.
\newblock


\bibitem[Hackenburg and Margetts(2024)]%
        {hackenburg2024evaluating}
\bibfield{author}{\bibinfo{person}{Kobi Hackenburg} {and} \bibinfo{person}{Helen Margetts}.} \bibinfo{year}{2024}\natexlab{}.
\newblock \showarticletitle{Evaluating the persuasive influence of political microtargeting with large language models}.
\newblock \bibinfo{journal}{\emph{Proceedings of the National Academy of Sciences}} \bibinfo{volume}{121}, \bibinfo{number}{24} (\bibinfo{year}{2024}), \bibinfo{pages}{e2403116121}.
\newblock


\bibitem[Hadfi et~al\mbox{.}(2023)]%
        {hadfi2023conversational}
\bibfield{author}{\bibinfo{person}{Rafik Hadfi}, \bibinfo{person}{Shun Okuhara}, \bibinfo{person}{Jawad Haqbeen}, \bibinfo{person}{Sofia Sahab}, \bibinfo{person}{Susumu Ohnuma}, {and} \bibinfo{person}{Takayuki Ito}.} \bibinfo{year}{2023}\natexlab{}.
\newblock \showarticletitle{Conversational agents enhance women's contribution in online debates}.
\newblock \bibinfo{journal}{\emph{Scientific Reports}} \bibinfo{volume}{13}, \bibinfo{number}{1} (\bibinfo{year}{2023}), \bibinfo{pages}{14534}.
\newblock


\bibitem[Hong et~al\mbox{.}(2023)]%
        {hong2023metagpt}
\bibfield{author}{\bibinfo{person}{Sirui Hong}, \bibinfo{person}{Xiawu Zheng}, \bibinfo{person}{Jonathan Chen}, \bibinfo{person}{Yuheng Cheng}, \bibinfo{person}{Jinlin Wang}, \bibinfo{person}{Ceyao Zhang}, \bibinfo{person}{Zili Wang}, \bibinfo{person}{Steven Ka~Shing Yau}, \bibinfo{person}{Zijuan Lin}, \bibinfo{person}{Liyang Zhou}, {et~al\mbox{.}}} \bibinfo{year}{2023}\natexlab{}.
\newblock \showarticletitle{Metagpt: Meta programming for multi-agent collaborative framework}.
\newblock \bibinfo{journal}{\emph{arXiv preprint arXiv:2308.00352}} (\bibinfo{year}{2023}).
\newblock


\bibitem[Hu et~al\mbox{.}(2018)]%
        {hu2018touch}
\bibfield{author}{\bibinfo{person}{Tianran Hu}, \bibinfo{person}{Anbang Xu}, \bibinfo{person}{Zhe Liu}, \bibinfo{person}{Quanzeng You}, \bibinfo{person}{Yufan Guo}, \bibinfo{person}{Vibha Sinha}, \bibinfo{person}{Jiebo Luo}, {and} \bibinfo{person}{Rama Akkiraju}.} \bibinfo{year}{2018}\natexlab{}.
\newblock \showarticletitle{Touch your heart: A tone-aware chatbot for customer care on social media}. In \bibinfo{booktitle}{\emph{Proceedings of the 2018 CHI conference on human factors in computing systems}}. \bibinfo{pages}{1--12}.
\newblock


\bibitem[Hui and Buchegger(2009)]%
        {hui2009groupthink}
\bibfield{author}{\bibinfo{person}{Pan Hui} {and} \bibinfo{person}{Sonja Buchegger}.} \bibinfo{year}{2009}\natexlab{}.
\newblock \showarticletitle{Groupthink and peer pressure: Social influence in online social network groups}. In \bibinfo{booktitle}{\emph{2009 International Conference on Advances in Social Network Analysis and Mining}}. IEEE, \bibinfo{pages}{53--59}.
\newblock


\bibitem[Isenberg(1986)]%
        {isenberg1986group}
\bibfield{author}{\bibinfo{person}{Daniel~J Isenberg}.} \bibinfo{year}{1986}\natexlab{}.
\newblock \showarticletitle{Group polarization: A critical review and meta-analysis.}
\newblock \bibinfo{journal}{\emph{Journal of personality and social psychology}} \bibinfo{volume}{50}, \bibinfo{number}{6} (\bibinfo{year}{1986}), \bibinfo{pages}{1141}.
\newblock


\bibitem[Jakesch et~al\mbox{.}(2023)]%
        {jakesch2023co}
\bibfield{author}{\bibinfo{person}{Maurice Jakesch}, \bibinfo{person}{Advait Bhat}, \bibinfo{person}{Daniel Buschek}, \bibinfo{person}{Lior Zalmanson}, {and} \bibinfo{person}{Mor Naaman}.} \bibinfo{year}{2023}\natexlab{}.
\newblock \showarticletitle{Co-writing with opinionated language models affects users’ views}. In \bibinfo{booktitle}{\emph{Proceedings of the 2023 CHI conference on human factors in computing systems}}. \bibinfo{pages}{1--15}.
\newblock


\bibitem[Jiang et~al\mbox{.}(2022)]%
        {jiang2022ai}
\bibfield{author}{\bibinfo{person}{Hua Jiang}, \bibinfo{person}{Yang Cheng}, \bibinfo{person}{Jeongwon Yang}, {and} \bibinfo{person}{Shanbing Gao}.} \bibinfo{year}{2022}\natexlab{}.
\newblock \showarticletitle{AI-powered chatbot communication with customers: Dialogic interactions, satisfaction, engagement, and customer behavior}.
\newblock \bibinfo{journal}{\emph{Computers in Human Behavior}}  \bibinfo{volume}{134} (\bibinfo{year}{2022}), \bibinfo{pages}{107329}.
\newblock


\bibitem[Jiang et~al\mbox{.}(2023)]%
        {jiang2023communitybots}
\bibfield{author}{\bibinfo{person}{Zhiqiu Jiang}, \bibinfo{person}{Mashrur Rashik}, \bibinfo{person}{Kunjal Panchal}, \bibinfo{person}{Mahmood Jasim}, \bibinfo{person}{Ali Sarvghad}, \bibinfo{person}{Pari Riahi}, \bibinfo{person}{Erica DeWitt}, \bibinfo{person}{Fey Thurber}, {and} \bibinfo{person}{Narges Mahyar}.} \bibinfo{year}{2023}\natexlab{}.
\newblock \showarticletitle{CommunityBots: creating and evaluating A multi-agent chatbot platform for public input elicitation}.
\newblock \bibinfo{journal}{\emph{Proceedings of the ACM on Human-Computer Interaction}} \bibinfo{volume}{7}, \bibinfo{number}{CSCW1} (\bibinfo{year}{2023}), \bibinfo{pages}{1--32}.
\newblock


\bibitem[Kim et~al\mbox{.}(2024)]%
        {kim2024engaged}
\bibfield{author}{\bibinfo{person}{Hanseob Kim}, \bibinfo{person}{Bin Han}, \bibinfo{person}{Jieun Kim}, \bibinfo{person}{Muhammad Firdaus~Syawaludin Lubis}, \bibinfo{person}{Gerard~Jounghyun Kim}, {and} \bibinfo{person}{Jae-In Hwang}.} \bibinfo{year}{2024}\natexlab{}.
\newblock \showarticletitle{Engaged and Affective Virtual Agents: Their Impact on Social Presence, Trustworthiness, and Decision-Making in the Group Discussion}. In \bibinfo{booktitle}{\emph{Proceedings of the CHI Conference on Human Factors in Computing Systems}}. \bibinfo{pages}{1--17}.
\newblock


\bibitem[Kim et~al\mbox{.}(2020)]%
        {kim2020bot}
\bibfield{author}{\bibinfo{person}{Soomin Kim}, \bibinfo{person}{Jinsu Eun}, \bibinfo{person}{Changhoon Oh}, \bibinfo{person}{Bongwon Suh}, {and} \bibinfo{person}{Joonhwan Lee}.} \bibinfo{year}{2020}\natexlab{}.
\newblock \showarticletitle{Bot in the bunch: Facilitating group chat discussion by improving efficiency and participation with a chatbot}. In \bibinfo{booktitle}{\emph{Proceedings of the 2020 CHI Conference on Human Factors in Computing Systems}}. \bibinfo{pages}{1--13}.
\newblock


\bibitem[Kim et~al\mbox{.}(2021)]%
        {kim2021moderator}
\bibfield{author}{\bibinfo{person}{Soomin Kim}, \bibinfo{person}{Jinsu Eun}, \bibinfo{person}{Joseph Seering}, {and} \bibinfo{person}{Joonhwan Lee}.} \bibinfo{year}{2021}\natexlab{}.
\newblock \showarticletitle{Moderator chatbot for deliberative discussion: Effects of discussion structure and discussant facilitation}.
\newblock \bibinfo{journal}{\emph{Proceedings of the ACM on Human-Computer Interaction}} \bibinfo{volume}{5}, \bibinfo{number}{CSCW1} (\bibinfo{year}{2021}), \bibinfo{pages}{1--26}.
\newblock


\bibitem[Kobus(2003)]%
        {kobus2003peers}
\bibfield{author}{\bibinfo{person}{Kimberly Kobus}.} \bibinfo{year}{2003}\natexlab{}.
\newblock \showarticletitle{Peers and adolescent smoking}.
\newblock \bibinfo{journal}{\emph{Addiction}}  \bibinfo{volume}{98} (\bibinfo{year}{2003}), \bibinfo{pages}{37--55}.
\newblock


\bibitem[Lee et~al\mbox{.}(2025)]%
        {lee2025beyond}
\bibfield{author}{\bibinfo{person}{Soohwan Lee}, \bibinfo{person}{Seoyeong Hwang}, {and} \bibinfo{person}{Kyungho Lee}.} \bibinfo{year}{2025}\natexlab{}.
\newblock \showarticletitle{Beyond Individual UX: Defining Group Experience (GX) as a New Paradigm for Group-centered AI}. In \bibinfo{booktitle}{\emph{Companion Publication of the 2025 ACM Designing Interactive Systems Conference}}. \bibinfo{pages}{357--362}.
\newblock


\bibitem[Li et~al\mbox{.}(2025a)]%
        {li2025we}
\bibfield{author}{\bibinfo{person}{Jingshu Li}, \bibinfo{person}{Tianqi Song}, \bibinfo{person}{Beichen Xue}, {and} \bibinfo{person}{Yi-Chieh Lee}.} \bibinfo{year}{2025}\natexlab{a}.
\newblock \showarticletitle{We Shape AI, and Thereafter AI Shape Us: Humans Align with AI through Social Influences}. In \bibinfo{booktitle}{\emph{ICLR 2025 Workshop on Bidirectional Human-AI Alignment}}.
\newblock


\bibitem[Li et~al\mbox{.}(2025b)]%
        {li2025confidence}
\bibfield{author}{\bibinfo{person}{Jingshu Li}, \bibinfo{person}{Yitian Yang}, \bibinfo{person}{Q~Vera Liao}, \bibinfo{person}{Junti Zhang}, {and} \bibinfo{person}{Yi-Chieh Lee}.} \bibinfo{year}{2025}\natexlab{b}.
\newblock \showarticletitle{As Confidence Aligns: Understanding the Effect of AI Confidence on Human Self-confidence in Human-AI Decision Making}. In \bibinfo{booktitle}{\emph{Proceedings of the 2025 CHI Conference on Human Factors in Computing Systems}}. \bibinfo{pages}{1--16}.
\newblock


\bibitem[Lieb and Goel(2024)]%
        {lieb2024student}
\bibfield{author}{\bibinfo{person}{Anna Lieb} {and} \bibinfo{person}{Toshali Goel}.} \bibinfo{year}{2024}\natexlab{}.
\newblock \showarticletitle{Student Interaction with NewtBot: An LLM-as-tutor Chatbot for Secondary Physics Education}. In \bibinfo{booktitle}{\emph{Extended Abstracts of the CHI Conference on Human Factors in Computing Systems}}. \bibinfo{pages}{1--8}.
\newblock


\bibitem[Light et~al\mbox{.}(2023)]%
        {light2023avalonbench}
\bibfield{author}{\bibinfo{person}{Jonathan Light}, \bibinfo{person}{Min Cai}, \bibinfo{person}{Sheng Shen}, {and} \bibinfo{person}{Ziniu Hu}.} \bibinfo{year}{2023}\natexlab{}.
\newblock \showarticletitle{AvalonBench: Evaluating LLMs Playing the Game of Avalon}. In \bibinfo{booktitle}{\emph{NeurIPS 2023 Foundation Models for Decision Making Workshop}}.
\newblock


\bibitem[Luger and Sellen(2016)]%
        {luger2016like}
\bibfield{author}{\bibinfo{person}{Ewa Luger} {and} \bibinfo{person}{Abigail Sellen}.} \bibinfo{year}{2016}\natexlab{}.
\newblock \showarticletitle{" Like Having a Really Bad PA" The Gulf between User Expectation and Experience of Conversational Agents}. In \bibinfo{booktitle}{\emph{Proceedings of the 2016 CHI conference on human factors in computing systems}}. \bibinfo{pages}{5286--5297}.
\newblock


\bibitem[Ma et~al\mbox{.}(2024)]%
        {ma2024beyond}
\bibfield{author}{\bibinfo{person}{Xiao Ma}, \bibinfo{person}{Swaroop Mishra}, \bibinfo{person}{Ariel Liu}, \bibinfo{person}{Sophie~Ying Su}, \bibinfo{person}{Jilin Chen}, \bibinfo{person}{Chinmay Kulkarni}, \bibinfo{person}{Heng-Tze Cheng}, \bibinfo{person}{Quoc Le}, {and} \bibinfo{person}{Ed Chi}.} \bibinfo{year}{2024}\natexlab{}.
\newblock \showarticletitle{Beyond chatbots: Explorellm for structured thoughts and personalized model responses}. In \bibinfo{booktitle}{\emph{Extended Abstracts of the CHI Conference on Human Factors in Computing Systems}}. \bibinfo{pages}{1--12}.
\newblock


\bibitem[McDonald and Crandall(2015)]%
        {mcdonald2015social}
\bibfield{author}{\bibinfo{person}{Rachel~I McDonald} {and} \bibinfo{person}{Christian~S Crandall}.} \bibinfo{year}{2015}\natexlab{}.
\newblock \showarticletitle{Social norms and social influence}.
\newblock \bibinfo{journal}{\emph{Current Opinion in Behavioral Sciences}}  \bibinfo{volume}{3} (\bibinfo{year}{2015}), \bibinfo{pages}{147--151}.
\newblock


\bibitem[Mehrabian and Stefl(1995)]%
        {mehrabian1995basic}
\bibfield{author}{\bibinfo{person}{Albert Mehrabian} {and} \bibinfo{person}{Catherine~A Stefl}.} \bibinfo{year}{1995}\natexlab{}.
\newblock \showarticletitle{Basic temperament components of loneliness, shyness, and conformity}.
\newblock \bibinfo{journal}{\emph{Social Behavior and Personality: an international journal}} \bibinfo{volume}{23}, \bibinfo{number}{3} (\bibinfo{year}{1995}), \bibinfo{pages}{253--263}.
\newblock


\bibitem[Myers and Lamm(1976)]%
        {myers1976group}
\bibfield{author}{\bibinfo{person}{David~G Myers} {and} \bibinfo{person}{Helmut Lamm}.} \bibinfo{year}{1976}\natexlab{}.
\newblock \showarticletitle{The group polarization phenomenon.}
\newblock \bibinfo{journal}{\emph{Psychological bulletin}} \bibinfo{volume}{83}, \bibinfo{number}{4} (\bibinfo{year}{1976}), \bibinfo{pages}{602}.
\newblock


\bibitem[Nass et~al\mbox{.}(1994)]%
        {nass1994computers}
\bibfield{author}{\bibinfo{person}{Clifford Nass}, \bibinfo{person}{Jonathan Steuer}, {and} \bibinfo{person}{Ellen~R Tauber}.} \bibinfo{year}{1994}\natexlab{}.
\newblock \showarticletitle{Computers are social actors}. In \bibinfo{booktitle}{\emph{Proceedings of the SIGCHI conference on Human factors in computing systems}}. \bibinfo{pages}{72--78}.
\newblock


\bibitem[Ning et~al\mbox{.}(2024)]%
        {ning2024cheatagent}
\bibfield{author}{\bibinfo{person}{Liang-bo Ning}, \bibinfo{person}{Shijie Wang}, \bibinfo{person}{Wenqi Fan}, \bibinfo{person}{Qing Li}, \bibinfo{person}{Xin Xu}, \bibinfo{person}{Hao Chen}, {and} \bibinfo{person}{Feiran Huang}.} \bibinfo{year}{2024}\natexlab{}.
\newblock \showarticletitle{Cheatagent: Attacking llm-empowered recommender systems via llm agent}. In \bibinfo{booktitle}{\emph{Proceedings of the 30th ACM SIGKDD Conference on Knowledge Discovery and Data Mining}}. \bibinfo{pages}{2284--2295}.
\newblock


\bibitem[Oostveen et~al\mbox{.}(1996)]%
        {oostveen1996social}
\bibfield{author}{\bibinfo{person}{Ton Oostveen}, \bibinfo{person}{Ronald Knibbe}, {and} \bibinfo{person}{Hein De~Vries}.} \bibinfo{year}{1996}\natexlab{}.
\newblock \showarticletitle{Social influences on young adults' alcohol consumption: norms, modeling, pressure, socializing, and conformity}.
\newblock \bibinfo{journal}{\emph{Addictive behaviors}} \bibinfo{volume}{21}, \bibinfo{number}{2} (\bibinfo{year}{1996}), \bibinfo{pages}{187--197}.
\newblock


\bibitem[Oyebode et~al\mbox{.}(2021)]%
        {oyebode2021tailoring}
\bibfield{author}{\bibinfo{person}{Oladapo Oyebode}, \bibinfo{person}{Chinenye Ndulue}, \bibinfo{person}{Dinesh Mulchandani}, \bibinfo{person}{Ashfaq A.~Zamil~Adib}, \bibinfo{person}{Mona Alhasani}, {and} \bibinfo{person}{Rita Orji}.} \bibinfo{year}{2021}\natexlab{}.
\newblock \showarticletitle{Tailoring persuasive and behaviour change systems based on stages of change and motivation}. In \bibinfo{booktitle}{\emph{Proceedings of the 2021 CHI conference on human factors in computing systems}}. \bibinfo{pages}{1--19}.
\newblock


\bibitem[Ozdemir et~al\mbox{.}(2023)]%
        {ozdemir2023human}
\bibfield{author}{\bibinfo{person}{Ozan Ozdemir}, \bibinfo{person}{Bora Kolfal}, \bibinfo{person}{Paul~R Messinger}, {and} \bibinfo{person}{Shaheer Rizvi}.} \bibinfo{year}{2023}\natexlab{}.
\newblock \showarticletitle{Human or virtual: How influencer type shapes brand attitudes}.
\newblock \bibinfo{journal}{\emph{Computers in Human Behavior}}  \bibinfo{volume}{145} (\bibinfo{year}{2023}), \bibinfo{pages}{107771}.
\newblock


\bibitem[Park et~al\mbox{.}(2023a)]%
        {park2023choicemates}
\bibfield{author}{\bibinfo{person}{Jeongeon Park}, \bibinfo{person}{Bryan Min}, \bibinfo{person}{Xiaojuan Ma}, {and} \bibinfo{person}{Juho Kim}.} \bibinfo{year}{2023}\natexlab{a}.
\newblock \showarticletitle{Choicemates: Supporting unfamiliar online decision-making with multi-agent conversational interactions}.
\newblock \bibinfo{journal}{\emph{arXiv preprint arXiv:2310.01331}} (\bibinfo{year}{2023}).
\newblock


\bibitem[Park et~al\mbox{.}(2023b)]%
        {park2023generative}
\bibfield{author}{\bibinfo{person}{Joon~Sung Park}, \bibinfo{person}{Joseph O'Brien}, \bibinfo{person}{Carrie~Jun Cai}, \bibinfo{person}{Meredith~Ringel Morris}, \bibinfo{person}{Percy Liang}, {and} \bibinfo{person}{Michael~S Bernstein}.} \bibinfo{year}{2023}\natexlab{b}.
\newblock \showarticletitle{Generative agents: Interactive simulacra of human behavior}. In \bibinfo{booktitle}{\emph{Proceedings of the 36th Annual ACM Symposium on User Interface Software and Technology}}. \bibinfo{pages}{1--22}.
\newblock


\bibitem[Park et~al\mbox{.}(2022)]%
        {park2022social}
\bibfield{author}{\bibinfo{person}{Joon~Sung Park}, \bibinfo{person}{Lindsay Popowski}, \bibinfo{person}{Carrie Cai}, \bibinfo{person}{Meredith~Ringel Morris}, \bibinfo{person}{Percy Liang}, {and} \bibinfo{person}{Michael~S Bernstein}.} \bibinfo{year}{2022}\natexlab{}.
\newblock \showarticletitle{Social simulacra: Creating populated prototypes for social computing systems}. In \bibinfo{booktitle}{\emph{Proceedings of the 35th Annual ACM Symposium on User Interface Software and Technology}}. \bibinfo{pages}{1--18}.
\newblock


\bibitem[Pataranutaporn et~al\mbox{.}(2023)]%
        {pataranutaporn2023influencing}
\bibfield{author}{\bibinfo{person}{Pat Pataranutaporn}, \bibinfo{person}{Ruby Liu}, \bibinfo{person}{Ed Finn}, {and} \bibinfo{person}{Pattie Maes}.} \bibinfo{year}{2023}\natexlab{}.
\newblock \showarticletitle{Influencing human--AI interaction by priming beliefs about AI can increase perceived trustworthiness, empathy and effectiveness}.
\newblock \bibinfo{journal}{\emph{Nature Machine Intelligence}} \bibinfo{volume}{5}, \bibinfo{number}{10} (\bibinfo{year}{2023}), \bibinfo{pages}{1076--1086}.
\newblock


\bibitem[Price et~al\mbox{.}(2006)]%
        {price2006normative}
\bibfield{author}{\bibinfo{person}{Vincent Price}, \bibinfo{person}{Lilach Nir}, {and} \bibinfo{person}{Joseph~N Cappella}.} \bibinfo{year}{2006}\natexlab{}.
\newblock \showarticletitle{Normative and informational influences in online political discussions}.
\newblock \bibinfo{journal}{\emph{Communication Theory}} \bibinfo{volume}{16}, \bibinfo{number}{1} (\bibinfo{year}{2006}), \bibinfo{pages}{47--74}.
\newblock


\bibitem[Qian et~al\mbox{.}(2023)]%
        {qian2023communicative}
\bibfield{author}{\bibinfo{person}{Chen Qian}, \bibinfo{person}{Xin Cong}, \bibinfo{person}{Cheng Yang}, \bibinfo{person}{Weize Chen}, \bibinfo{person}{Yusheng Su}, \bibinfo{person}{Juyuan Xu}, \bibinfo{person}{Zhiyuan Liu}, {and} \bibinfo{person}{Maosong Sun}.} \bibinfo{year}{2023}\natexlab{}.
\newblock \showarticletitle{Communicative agents for software development}.
\newblock \bibinfo{journal}{\emph{arXiv preprint arXiv:2307.07924}} (\bibinfo{year}{2023}).
\newblock


\bibitem[{Reuters}(2024)]%
        {reuters2024openai}
\bibfield{author}{\bibinfo{person}{{Reuters}}.} \bibinfo{year}{2024}\natexlab{}.
\newblock \bibinfo{title}{OpenAI says ChatGPT's weekly users have grown to 200 million}.
\newblock \bibinfo{howpublished}{\url{https://www.reuters.com/technology/artificial-intelligence/openai-says-chatgpts-weekly-users-have-grown-200-million-2024-08-29/}}.
\newblock
\newblock
\shownote{Accessed: 2024-10-28}.


\bibitem[Rezwana and Maher(2023)]%
        {rezwana2023designing}
\bibfield{author}{\bibinfo{person}{Jeba Rezwana} {and} \bibinfo{person}{Mary~Lou Maher}.} \bibinfo{year}{2023}\natexlab{}.
\newblock \showarticletitle{Designing creative AI partners with COFI: A framework for modeling interaction in human-AI co-creative systems}.
\newblock \bibinfo{journal}{\emph{ACM Transactions on Computer-Human Interaction}} \bibinfo{volume}{30}, \bibinfo{number}{5} (\bibinfo{year}{2023}), \bibinfo{pages}{1--28}.
\newblock


\bibitem[Richards et~al\mbox{.}(2022)]%
        {richards2022reducing}
\bibfield{author}{\bibinfo{person}{Adam~S Richards}, \bibinfo{person}{Elena Bessarabova}, \bibinfo{person}{John~A Banas}, {and} \bibinfo{person}{Daniel~R Bernard}.} \bibinfo{year}{2022}\natexlab{}.
\newblock \showarticletitle{Reducing psychological reactance to health promotion messages: Comparing preemptive and postscript mitigation strategies}.
\newblock \bibinfo{journal}{\emph{Health Communication}} \bibinfo{volume}{37}, \bibinfo{number}{3} (\bibinfo{year}{2022}), \bibinfo{pages}{366--374}.
\newblock


\bibitem[Richters and Waters(1991)]%
        {richters1991attachment}
\bibfield{author}{\bibinfo{person}{John~E Richters} {and} \bibinfo{person}{Everett Waters}.} \bibinfo{year}{1991}\natexlab{}.
\newblock \showarticletitle{Attachment and socialization: The positive side of social influence}.
\newblock In \bibinfo{booktitle}{\emph{Social influences and socialization in infancy}}. \bibinfo{publisher}{Springer}, \bibinfo{pages}{185--213}.
\newblock


\bibitem[Rosenberg and Siegel(2018)]%
        {rosenberg201850}
\bibfield{author}{\bibinfo{person}{Benjamin~D Rosenberg} {and} \bibinfo{person}{Jason~T Siegel}.} \bibinfo{year}{2018}\natexlab{}.
\newblock \showarticletitle{A 50-year review of psychological reactance theory: Do not read this article.}
\newblock \bibinfo{journal}{\emph{Motivation Science}} \bibinfo{volume}{4}, \bibinfo{number}{4} (\bibinfo{year}{2018}), \bibinfo{pages}{281}.
\newblock


\bibitem[Roy et~al\mbox{.}(2024)]%
        {roy2024exploring}
\bibfield{author}{\bibinfo{person}{Devjeet Roy}, \bibinfo{person}{Xuchao Zhang}, \bibinfo{person}{Rashi Bhave}, \bibinfo{person}{Chetan Bansal}, \bibinfo{person}{Pedro Las-Casas}, \bibinfo{person}{Rodrigo Fonseca}, {and} \bibinfo{person}{Saravan Rajmohan}.} \bibinfo{year}{2024}\natexlab{}.
\newblock \showarticletitle{Exploring llm-based agents for root cause analysis}. In \bibinfo{booktitle}{\emph{Companion Proceedings of the 32nd ACM International Conference on the Foundations of Software Engineering}}. \bibinfo{pages}{208--219}.
\newblock


\bibitem[Rutkowski et~al\mbox{.}(1983)]%
        {rutkowski1983group}
\bibfield{author}{\bibinfo{person}{Gregory~K Rutkowski}, \bibinfo{person}{Charles~L Gruder}, {and} \bibinfo{person}{Daniel Romer}.} \bibinfo{year}{1983}\natexlab{}.
\newblock \showarticletitle{Group cohesiveness, social norms, and bystander intervention.}
\newblock \bibinfo{journal}{\emph{Journal of Personality and Social Psychology}} \bibinfo{volume}{44}, \bibinfo{number}{3} (\bibinfo{year}{1983}), \bibinfo{pages}{545}.
\newblock


\bibitem[Salganik et~al\mbox{.}(2006)]%
        {salganik2006experimental}
\bibfield{author}{\bibinfo{person}{Matthew~J Salganik}, \bibinfo{person}{Peter~Sheridan Dodds}, {and} \bibinfo{person}{Duncan~J Watts}.} \bibinfo{year}{2006}\natexlab{}.
\newblock \showarticletitle{Experimental study of inequality and unpredictability in an artificial cultural market}.
\newblock \bibinfo{journal}{\emph{science}} \bibinfo{volume}{311}, \bibinfo{number}{5762} (\bibinfo{year}{2006}), \bibinfo{pages}{854--856}.
\newblock


\bibitem[Sas et~al\mbox{.}(2023)]%
        {sas2023informing}
\bibfield{author}{\bibinfo{person}{Martin Sas}, \bibinfo{person}{Maarten Denoo}, {and} \bibinfo{person}{Jan~Tobias M{\"u}hlberg}.} \bibinfo{year}{2023}\natexlab{}.
\newblock \showarticletitle{Informing Children about Privacy: A Review and Assessment of Age-Appropriate Information Designs in Kids-Oriented F2P Video Games}.
\newblock \bibinfo{journal}{\emph{Proceedings of the ACM on Human-Computer Interaction}} \bibinfo{volume}{7}, \bibinfo{number}{CHI PLAY} (\bibinfo{year}{2023}), \bibinfo{pages}{425--463}.
\newblock


\bibitem[Sebastian et~al\mbox{.}(2023)]%
        {sebastian2023exploring}
\bibfield{author}{\bibinfo{person}{Glorin Sebastian}, \bibinfo{person}{Rose Sebastian~Sr}, {et~al\mbox{.}}} \bibinfo{year}{2023}\natexlab{}.
\newblock \showarticletitle{Exploring ethical implications of ChatGPT and other AI chatbots and regulation of disinformation propagation}.
\newblock \bibinfo{journal}{\emph{Exploring Ethical Implications of ChatGPT and Other AI Chatbots and Regulation of Disinformation Propagation (May 29, 2023)}} (\bibinfo{year}{2023}).
\newblock


\bibitem[Sheehan et~al\mbox{.}(2020)]%
        {sheehan2020customer}
\bibfield{author}{\bibinfo{person}{Ben Sheehan}, \bibinfo{person}{Hyun~Seung Jin}, {and} \bibinfo{person}{Udo Gottlieb}.} \bibinfo{year}{2020}\natexlab{}.
\newblock \showarticletitle{Customer service chatbots: Anthropomorphism and adoption}.
\newblock \bibinfo{journal}{\emph{Journal of Business Research}}  \bibinfo{volume}{115} (\bibinfo{year}{2020}), \bibinfo{pages}{14--24}.
\newblock


\bibitem[Shepherd et~al\mbox{.}(2011)]%
        {shepherd2011susceptible}
\bibfield{author}{\bibinfo{person}{Jennifer~L Shepherd}, \bibinfo{person}{David~J Lane}, \bibinfo{person}{Ryan~L Tapscott}, {and} \bibinfo{person}{Douglas~A Gentile}.} \bibinfo{year}{2011}\natexlab{}.
\newblock \showarticletitle{Susceptible to Social Influence: Risky “Driving” in Response to Peer Pressure 1}.
\newblock \bibinfo{journal}{\emph{Journal of Applied Social Psychology}} \bibinfo{volume}{41}, \bibinfo{number}{4} (\bibinfo{year}{2011}), \bibinfo{pages}{773--797}.
\newblock


\bibitem[Shi et~al\mbox{.}(2020)]%
        {shi2020effects}
\bibfield{author}{\bibinfo{person}{Weiyan Shi}, \bibinfo{person}{Xuewei Wang}, \bibinfo{person}{Yoo~Jung Oh}, \bibinfo{person}{Jingwen Zhang}, \bibinfo{person}{Saurav Sahay}, {and} \bibinfo{person}{Zhou Yu}.} \bibinfo{year}{2020}\natexlab{}.
\newblock \showarticletitle{Effects of persuasive dialogues: testing bot identities and inquiry strategies}. In \bibinfo{booktitle}{\emph{Proceedings of the 2020 CHI Conference on Human Factors in Computing Systems}}. \bibinfo{pages}{1--13}.
\newblock


\bibitem[Sidji et~al\mbox{.}(2024)]%
        {sidji2024human}
\bibfield{author}{\bibinfo{person}{Matthew Sidji}, \bibinfo{person}{Wally Smith}, {and} \bibinfo{person}{Melissa~J Rogerson}.} \bibinfo{year}{2024}\natexlab{}.
\newblock \showarticletitle{Human-AI Collaboration in Cooperative Games: A Study of Playing Codenames with an LLM Assistant}.
\newblock \bibinfo{journal}{\emph{Proceedings of the ACM on Human-Computer Interaction}} \bibinfo{volume}{8}, \bibinfo{number}{CHI PLAY} (\bibinfo{year}{2024}), \bibinfo{pages}{1--25}.
\newblock


\bibitem[Skalski and Tamborini(2007)]%
        {skalski2007role}
\bibfield{author}{\bibinfo{person}{Paul Skalski} {and} \bibinfo{person}{Ron Tamborini}.} \bibinfo{year}{2007}\natexlab{}.
\newblock \showarticletitle{The role of social presence in interactive agent-based persuasion}.
\newblock \bibinfo{journal}{\emph{Media psychology}} \bibinfo{volume}{10}, \bibinfo{number}{3} (\bibinfo{year}{2007}), \bibinfo{pages}{385--413}.
\newblock


\bibitem[Smoliar(1991)]%
        {minsky1991society}
\bibfield{author}{\bibinfo{person}{Stephen~W. Smoliar}.} \bibinfo{year}{1991}\natexlab{}.
\newblock \showarticletitle{The society of mind: Marvin Minsky}.
\newblock \bibinfo{journal}{\emph{Artificial Intelligence}} \bibinfo{volume}{48}, \bibinfo{number}{3} (\bibinfo{year}{1991}), \bibinfo{pages}{349--370}.
\newblock
\showISSN{0004-3702}
\href{https://doi.org/10.1016/0004-3702(91)90035-I}{doi:\nolinkurl{10.1016/0004-3702(91)90035-I}}


\bibitem[Snow et~al\mbox{.}(2020)]%
        {snow2020rent}
\bibfield{author}{\bibinfo{person}{Stephen Snow}, \bibinfo{person}{Toby Guinea}, \bibinfo{person}{Alexander Balson}, \bibinfo{person}{Awais~Hameed Khan}, \bibinfo{person}{Mashhuda Glencross}, {and} \bibinfo{person}{Neil Horrocks}.} \bibinfo{year}{2020}\natexlab{}.
\newblock \showarticletitle{Rent-a-Watt: Rethinking energy use feedback}. In \bibinfo{booktitle}{\emph{Proceedings of the 32nd Australian Conference on Human-Computer Interaction}}. \bibinfo{pages}{736--741}.
\newblock


\bibitem[Snow et~al\mbox{.}(2021)]%
        {snow2021neighbourhood}
\bibfield{author}{\bibinfo{person}{Stephen Snow}, \bibinfo{person}{Awais~Hameed Khan}, \bibinfo{person}{Mashhuda Glencross}, {and} \bibinfo{person}{Neil Horrocks}.} \bibinfo{year}{2021}\natexlab{}.
\newblock \showarticletitle{Neighbourhood Wattch: Using Speculative Design to Explore Values Around Curtailment and Consent in Household Energy Interactions}. In \bibinfo{booktitle}{\emph{Proceedings of the 2021 CHI Conference on Human Factors in Computing Systems}}. \bibinfo{pages}{1--12}.
\newblock


\bibitem[Song and Shin(2024)]%
        {song2024uncanny}
\bibfield{author}{\bibinfo{person}{Stephen~Wonchul Song} {and} \bibinfo{person}{Mincheol Shin}.} \bibinfo{year}{2024}\natexlab{}.
\newblock \showarticletitle{Uncanny valley effects on chatbot trust, purchase intention, and adoption intention in the context of e-commerce: The moderating role of avatar familiarity}.
\newblock \bibinfo{journal}{\emph{International Journal of Human--Computer Interaction}} \bibinfo{volume}{40}, \bibinfo{number}{2} (\bibinfo{year}{2024}), \bibinfo{pages}{441--456}.
\newblock


\bibitem[Song et~al\mbox{.}(2025)]%
        {song2025greater}
\bibfield{author}{\bibinfo{person}{Tianqi Song}, \bibinfo{person}{Yugin Tan}, \bibinfo{person}{Zicheng Zhu}, \bibinfo{person}{Yibin Feng}, {and} \bibinfo{person}{Yi-Chieh Lee}.} \bibinfo{year}{2025}\natexlab{}.
\newblock \showarticletitle{Greater than the Sum of its Parts: Exploring Social Influence of Multi-Agents}. In \bibinfo{booktitle}{\emph{Proceedings of the Extended Abstracts of the CHI Conference on Human Factors in Computing Systems}}. \bibinfo{pages}{1--11}.
\newblock


\bibitem[Spartz et~al\mbox{.}(2017)]%
        {spartz2017youtube}
\bibfield{author}{\bibinfo{person}{James~T Spartz}, \bibinfo{person}{Leona Yi-Fan Su}, \bibinfo{person}{Robert Griffin}, \bibinfo{person}{Dominique Brossard}, {and} \bibinfo{person}{Sharon Dunwoody}.} \bibinfo{year}{2017}\natexlab{}.
\newblock \showarticletitle{YouTube, social norms and perceived salience of climate change in the American mind}.
\newblock \bibinfo{journal}{\emph{Environmental Communication}} \bibinfo{volume}{11}, \bibinfo{number}{1} (\bibinfo{year}{2017}), \bibinfo{pages}{1--16}.
\newblock


\bibitem[Spears(2021)]%
        {spears2021social}
\bibfield{author}{\bibinfo{person}{Russell Spears}.} \bibinfo{year}{2021}\natexlab{}.
\newblock \showarticletitle{Social influence and group identity}.
\newblock \bibinfo{journal}{\emph{Annual review of psychology}} \bibinfo{volume}{72}, \bibinfo{number}{1} (\bibinfo{year}{2021}), \bibinfo{pages}{367--390}.
\newblock


\bibitem[Steinberg and Monahan(2007)]%
        {steinberg2007age}
\bibfield{author}{\bibinfo{person}{Laurence Steinberg} {and} \bibinfo{person}{Kathryn~C Monahan}.} \bibinfo{year}{2007}\natexlab{}.
\newblock \showarticletitle{Age differences in resistance to peer influence.}
\newblock \bibinfo{journal}{\emph{Developmental psychology}} \bibinfo{volume}{43}, \bibinfo{number}{6} (\bibinfo{year}{2007}), \bibinfo{pages}{1531}.
\newblock


\bibitem[Tan and Liew(2022)]%
        {tan2022multi}
\bibfield{author}{\bibinfo{person}{Su-Mae Tan} {and} \bibinfo{person}{Tze~Wei Liew}.} \bibinfo{year}{2022}\natexlab{}.
\newblock \showarticletitle{Multi-Chatbot or Single-Chatbot? The Effects of M-Commerce Chatbot Interface on Source Credibility, Social Presence, Trust, and Purchase Intention}.
\newblock \bibinfo{journal}{\emph{Human Behavior and Emerging Technologies}} \bibinfo{volume}{2022}, \bibinfo{number}{1} (\bibinfo{year}{2022}), \bibinfo{pages}{2501538}.
\newblock


\bibitem[Tanprasert et~al\mbox{.}(2024)]%
        {tanprasert2024debate}
\bibfield{author}{\bibinfo{person}{Thitaree Tanprasert}, \bibinfo{person}{Sidney~S Fels}, \bibinfo{person}{Luanne Sinnamon}, {and} \bibinfo{person}{Dongwook Yoon}.} \bibinfo{year}{2024}\natexlab{}.
\newblock \showarticletitle{Debate Chatbots to Facilitate Critical Thinking on YouTube: Social Identity and Conversational Style Make A Difference}. In \bibinfo{booktitle}{\emph{Proceedings of the CHI Conference on Human Factors in Computing Systems}}. \bibinfo{pages}{1--24}.
\newblock


\bibitem[Teo et~al\mbox{.}(2019)]%
        {teo2019marketing}
\bibfield{author}{\bibinfo{person}{Li~Xin Teo}, \bibinfo{person}{Ho~Keat Leng}, {and} \bibinfo{person}{Yi~Xian~Philip Phua}.} \bibinfo{year}{2019}\natexlab{}.
\newblock \showarticletitle{Marketing on Instagram: Social influence and image quality on perception of quality and purchase intention}.
\newblock \bibinfo{journal}{\emph{International Journal of Sports Marketing and Sponsorship}} \bibinfo{volume}{20}, \bibinfo{number}{2} (\bibinfo{year}{2019}), \bibinfo{pages}{321--332}.
\newblock


\bibitem[Tian et~al\mbox{.}(2021)]%
        {tian2021let}
\bibfield{author}{\bibinfo{person}{Xiaoyi Tian}, \bibinfo{person}{Zak Risha}, \bibinfo{person}{Ishrat Ahmed}, \bibinfo{person}{Arun~Balajiee Lekshmi~Narayanan}, {and} \bibinfo{person}{Jacob Biehl}.} \bibinfo{year}{2021}\natexlab{}.
\newblock \showarticletitle{Let's talk it out: A chatbot for effective study habit behavioral change}.
\newblock \bibinfo{journal}{\emph{Proceedings of the ACM on Human-Computer Interaction}} \bibinfo{volume}{5}, \bibinfo{number}{CSCW1} (\bibinfo{year}{2021}), \bibinfo{pages}{1--32}.
\newblock


\bibitem[Trost et~al\mbox{.}(1992)]%
        {trost1992minority}
\bibfield{author}{\bibinfo{person}{Melanie~R Trost}, \bibinfo{person}{Anne Maass}, {and} \bibinfo{person}{Douglas~T Kenrick}.} \bibinfo{year}{1992}\natexlab{}.
\newblock \showarticletitle{Minority influence: Personal relevance biases cognitive processes and reverses private acceptance}.
\newblock \bibinfo{journal}{\emph{Journal of Experimental Social Psychology}} \bibinfo{volume}{28}, \bibinfo{number}{3} (\bibinfo{year}{1992}), \bibinfo{pages}{234--254}.
\newblock


\bibitem[Vossen et~al\mbox{.}(2009)]%
        {vossen2009social}
\bibfield{author}{\bibinfo{person}{Suzanne Vossen}, \bibinfo{person}{Jaap Ham}, {and} \bibinfo{person}{Cees Midden}.} \bibinfo{year}{2009}\natexlab{}.
\newblock \showarticletitle{Social influence of a persuasive agent: the role of agent embodiment and evaluative feedback}. In \bibinfo{booktitle}{\emph{Proceedings of the 4th International Conference on Persuasive Technology}}. \bibinfo{pages}{1--7}.
\newblock


\bibitem[Wang et~al\mbox{.}(2016)]%
        {wang2016influence}
\bibfield{author}{\bibinfo{person}{Dawei Wang}, \bibinfo{person}{Liping Zhu}, \bibinfo{person}{Phil Maguire}, \bibinfo{person}{Yixin Liu}, \bibinfo{person}{Kaiyuan Pang}, \bibinfo{person}{Zhenying Li}, {and} \bibinfo{person}{Yixin Hu}.} \bibinfo{year}{2016}\natexlab{}.
\newblock \showarticletitle{The influence of social comparison and peer group size on risky decision-making}.
\newblock \bibinfo{journal}{\emph{Frontiers in psychology}}  \bibinfo{volume}{7} (\bibinfo{year}{2016}), \bibinfo{pages}{1232}.
\newblock


\bibitem[Wang et~al\mbox{.}(2024b)]%
        {wang2024survey}
\bibfield{author}{\bibinfo{person}{Lei Wang}, \bibinfo{person}{Chen Ma}, \bibinfo{person}{Xueyang Feng}, \bibinfo{person}{Zeyu Zhang}, \bibinfo{person}{Hao Yang}, \bibinfo{person}{Jingsen Zhang}, \bibinfo{person}{Zhiyuan Chen}, \bibinfo{person}{Jiakai Tang}, \bibinfo{person}{Xu Chen}, \bibinfo{person}{Yankai Lin}, {et~al\mbox{.}}} \bibinfo{year}{2024}\natexlab{b}.
\newblock \showarticletitle{A survey on large language model based autonomous agents}.
\newblock \bibinfo{journal}{\emph{Frontiers of Computer Science}} \bibinfo{volume}{18}, \bibinfo{number}{6} (\bibinfo{year}{2024}), \bibinfo{pages}{186345}.
\newblock


\bibitem[Wang et~al\mbox{.}(2024a)]%
        {wang2024executable}
\bibfield{author}{\bibinfo{person}{Xingyao Wang}, \bibinfo{person}{Yangyi Chen}, \bibinfo{person}{Lifan Yuan}, \bibinfo{person}{Yizhe Zhang}, \bibinfo{person}{Yunzhu Li}, \bibinfo{person}{Hao Peng}, {and} \bibinfo{person}{Heng Ji}.} \bibinfo{year}{2024}\natexlab{a}.
\newblock \showarticletitle{Executable Code Actions Elicit Better LLM Agents}. In \bibinfo{booktitle}{\emph{Forty-first International Conference on Machine Learning}}.
\newblock


\bibitem[Wilder(1977)]%
        {wilder1977perception}
\bibfield{author}{\bibinfo{person}{David~A Wilder}.} \bibinfo{year}{1977}\natexlab{}.
\newblock \showarticletitle{Perception of groups, size of opposition, and social influence}.
\newblock \bibinfo{journal}{\emph{Journal of Experimental Social Psychology}} \bibinfo{volume}{13}, \bibinfo{number}{3} (\bibinfo{year}{1977}), \bibinfo{pages}{253--268}.
\newblock


\bibitem[Wischnewski et~al\mbox{.}(2024)]%
        {wischnewski2024agree}
\bibfield{author}{\bibinfo{person}{Magdalena Wischnewski}, \bibinfo{person}{Thao Ngo}, \bibinfo{person}{Rebecca Bernemann}, \bibinfo{person}{Martin Jansen}, {and} \bibinfo{person}{Nicole Kr{\"a}mer}.} \bibinfo{year}{2024}\natexlab{}.
\newblock \showarticletitle{“I agree with you, bot!” How users (dis) engage with social bots on Twitter}.
\newblock \bibinfo{journal}{\emph{New Media \& Society}} \bibinfo{volume}{26}, \bibinfo{number}{3} (\bibinfo{year}{2024}), \bibinfo{pages}{1505--1526}.
\newblock


\bibitem[Xiong et~al\mbox{.}(2023)]%
        {xiong2023examining}
\bibfield{author}{\bibinfo{person}{Kai Xiong}, \bibinfo{person}{Xiao Ding}, \bibinfo{person}{Yixin Cao}, \bibinfo{person}{Ting Liu}, {and} \bibinfo{person}{Bing Qin}.} \bibinfo{year}{2023}\natexlab{}.
\newblock \showarticletitle{Examining inter-consistency of large language models collaboration: An in-depth analysis via debate}. In \bibinfo{booktitle}{\emph{The 2023 Conference on Empirical Methods in Natural Language Processing}}.
\newblock


\bibitem[Xu et~al\mbox{.}(2017)]%
        {xu2017new}
\bibfield{author}{\bibinfo{person}{Anbang Xu}, \bibinfo{person}{Zhe Liu}, \bibinfo{person}{Yufan Guo}, \bibinfo{person}{Vibha Sinha}, {and} \bibinfo{person}{Rama Akkiraju}.} \bibinfo{year}{2017}\natexlab{}.
\newblock \showarticletitle{A new chatbot for customer service on social media}. In \bibinfo{booktitle}{\emph{Proceedings of the 2017 CHI conference on human factors in computing systems}}. \bibinfo{pages}{3506--3510}.
\newblock


\bibitem[Yang et~al\mbox{.}(2025)]%
        {yang2025understanding}
\bibfield{author}{\bibinfo{person}{Yitian Yang}, \bibinfo{person}{Yugin Tan}, \bibinfo{person}{Yang~Chen Lin}, \bibinfo{person}{Jung-Tai King}, \bibinfo{person}{Zihan Liu}, {and} \bibinfo{person}{Yi-Chieh Lee}.} \bibinfo{year}{2025}\natexlab{}.
\newblock \showarticletitle{Understanding How Psychological Distance Influences User Preferences in Conversational versus Web Search}. In \bibinfo{booktitle}{\emph{Proceedings of the 2025 CHI Conference on Human Factors in Computing Systems}}. \bibinfo{pages}{1--18}.
\newblock


\bibitem[Yeo et~al\mbox{.}(2024)]%
        {yeo2024help}
\bibfield{author}{\bibinfo{person}{ShunYi Yeo}, \bibinfo{person}{Gionnieve Lim}, \bibinfo{person}{Jie Gao}, \bibinfo{person}{Weiyu Zhang}, {and} \bibinfo{person}{Simon~Tangi Perrault}.} \bibinfo{year}{2024}\natexlab{}.
\newblock \showarticletitle{Help Me Reflect: Leveraging Self-Reflection Interface Nudges to Enhance Deliberativeness on Online Deliberation Platforms}. In \bibinfo{booktitle}{\emph{Proceedings of the CHI Conference on Human Factors in Computing Systems}}. \bibinfo{pages}{1--32}.
\newblock


\bibitem[Yu et~al\mbox{.}(2024)]%
        {yu2024finmem}
\bibfield{author}{\bibinfo{person}{Yangyang Yu}, \bibinfo{person}{Haohang Li}, \bibinfo{person}{Zhi Chen}, \bibinfo{person}{Yuechen Jiang}, \bibinfo{person}{Yang Li}, \bibinfo{person}{Denghui Zhang}, \bibinfo{person}{Rong Liu}, \bibinfo{person}{Jordan~W Suchow}, {and} \bibinfo{person}{Khaldoun Khashanah}.} \bibinfo{year}{2024}\natexlab{}.
\newblock \showarticletitle{FinMem: A performance-enhanced LLM trading agent with layered memory and character design}. In \bibinfo{booktitle}{\emph{Proceedings of the AAAI Symposium Series}}, Vol.~\bibinfo{volume}{3}. \bibinfo{pages}{595--597}.
\newblock


\bibitem[Zhang et~al\mbox{.}(2020)]%
        {zhang2020artificial}
\bibfield{author}{\bibinfo{person}{Jingwen Zhang}, \bibinfo{person}{Yoo~Jung Oh}, \bibinfo{person}{Patrick Lange}, \bibinfo{person}{Zhou Yu}, {and} \bibinfo{person}{Yoshimi Fukuoka}.} \bibinfo{year}{2020}\natexlab{}.
\newblock \showarticletitle{Artificial intelligence chatbot behavior change model for designing artificial intelligence chatbots to promote physical activity and a healthy diet}.
\newblock \bibinfo{journal}{\emph{Journal of medical Internet research}} \bibinfo{volume}{22}, \bibinfo{number}{9} (\bibinfo{year}{2020}), \bibinfo{pages}{e22845}.
\newblock


\bibitem[Zhang et~al\mbox{.}(2015)]%
        {zhang2015leveraging}
\bibfield{author}{\bibinfo{person}{Jun Zhang}, \bibinfo{person}{Liping Tong}, \bibinfo{person}{Peter~J Lamberson}, \bibinfo{person}{Ram{\'o}n~A Durazo-Arvizu}, \bibinfo{person}{A Luke}, {and} \bibinfo{person}{David~A Shoham}.} \bibinfo{year}{2015}\natexlab{}.
\newblock \showarticletitle{Leveraging social influence to address overweight and obesity using agent-based models: the role of adolescent social networks}.
\newblock \bibinfo{journal}{\emph{Social science \& medicine}}  \bibinfo{volume}{125} (\bibinfo{year}{2015}), \bibinfo{pages}{203--213}.
\newblock


\bibitem[Zhang et~al\mbox{.}(2024)]%
        {zhang2024see}
\bibfield{author}{\bibinfo{person}{Yu Zhang}, \bibinfo{person}{Jingwei Sun}, \bibinfo{person}{Li Feng}, \bibinfo{person}{Cen Yao}, \bibinfo{person}{Mingming Fan}, \bibinfo{person}{Liuxin Zhang}, \bibinfo{person}{Qianying Wang}, \bibinfo{person}{Xin Geng}, {and} \bibinfo{person}{Yong Rui}.} \bibinfo{year}{2024}\natexlab{}.
\newblock \showarticletitle{See Widely, Think Wisely: Toward Designing a Generative Multi-agent System to Burst Filter Bubbles}. In \bibinfo{booktitle}{\emph{Proceedings of the CHI Conference on Human Factors in Computing Systems}}. \bibinfo{pages}{1--24}.
\newblock


\bibitem[Zhu et~al\mbox{.}(2012)]%
        {zhu2012switch}
\bibfield{author}{\bibinfo{person}{Haiyi Zhu}, \bibinfo{person}{Bernardo Huberman}, {and} \bibinfo{person}{Yarun Luon}.} \bibinfo{year}{2012}\natexlab{}.
\newblock \showarticletitle{To switch or not to switch: understanding social influence in online choices}. In \bibinfo{booktitle}{\emph{Proceedings of the SIGCHI conference on human factors in computing systems}}. \bibinfo{pages}{2257--2266}.
\newblock


\bibitem[Zhu et~al\mbox{.}(2025)]%
        {zhu2025benefits}
\bibfield{author}{\bibinfo{person}{Zicheng Zhu}, \bibinfo{person}{Yugin Tan}, \bibinfo{person}{Naomi Yamashita}, \bibinfo{person}{Yi-Chieh Lee}, {and} \bibinfo{person}{Renwen Zhang}.} \bibinfo{year}{2025}\natexlab{}.
\newblock \showarticletitle{The Benefits of Prosociality towards AI Agents: Examining the Effects of Helping AI Agents on Human Well-Being}. In \bibinfo{booktitle}{\emph{Proceedings of the 2025 CHI Conference on Human Factors in Computing Systems}}. \bibinfo{pages}{1--18}.
\newblock


\end{thebibliography}

\appendix

\section{Appendix}

\subsection{General Impressions}
\label{app:impressions}

\subsubsection{Quantitative}

{\small
\begin{table*}
    \centering
    \caption{Summary of Users' General Impressions Across Experimental Conditions}
    \label{tab:general-impression}
    \begin{tabularx}{\textwidth}{l l *{6}{>{\centering\arraybackslash}X} c c}
    \toprule
        \textbf{Item} & \textbf{Scenario} & \multicolumn{2}{c}{\textbf{1 Agent}} & \multicolumn{2}{c}{\textbf{3 Agents}} & \multicolumn{2}{c}{\textbf{5 Agents}} & H/F & p \\
        ~ & ~ & Mean & STD & Mean & STD & Mean & STD & ~ & ~ \\
    \midrule

    \multicolumn{10}{l}{\textit{Understanding}} \\
        ~ & Topic 1 & 5.35 & 1.84 & 5.45 & 1.06 & 5.30 & 1.37 & 1.28 & 0.53 \\
        ~ & Topic 2 & 5.35 & 1.72 & 5.48 & 1.50 & 5.67 & 1.15 & 0.07 & 0.97 \\
        ~ & Same     & 5.48 & 1.61 & 5.73 & 1.21 & 5.87 & 1.04 & 0.52 & 0.77 \\
        ~ & Different& 5.23 & 1.93 & 5.21 & 1.34 & 5.10 & 1.37 & 2.29 & 0.32 \\

    \midrule
    \multicolumn{10}{l}{\textit{Expertise}} \\
        ~ & Topic 1 & 5.29 & 1.83 & 5.48 & 1.15 & 5.30 & 1.29 & 0.87 & 0.65 \\
        ~ & Topic 2 & 5.13 & 1.48 & 5.21 & 1.36 & 5.33 & 1.24 & 0.09 & 0.96 \\
        ~ & Same     & 5.45 & 1.52 & 5.61 & 1.20 & 5.67 & 1.21 & 0.08 & 0.96 \\
        ~ & Different& 4.97 & 1.76 & 5.09 & 1.28 & 4.97 & 1.22 & 0.69 & 0.71 \\

    \midrule
    \multicolumn{10}{l}{\textit{Balanced}} \\
        ~ & Topic 1 & 5.16 & 2.07 & 5.24 & 1.41 & 5.50 & 1.04 & 0.90 & 0.64 \\
        ~ & Topic 2 & 5.06 & 1.86 & 5.06 & 1.50 & 5.60 & 1.22 & 1.92 & 0.38 \\
        ~ & Same     & 5.48 & 1.65 & 5.61 & 1.22 & 5.80 & 1.10 & 0.40 & 0.82 \\
        ~ & Different& 4.74 & 2.18 & 4.70 & 1.53 & 5.30 & 1.12 & 1.25 & 0.29 \\

    \midrule
    \multicolumn{10}{l}{\textit{Inspired}} \\
        ~ & Topic 1 & 3.68 & 2.06 & 4.18 & 2.16 & 3.70 & 1.90 & 1.32 & 0.52 \\
        ~ & Topic 2 & 3.10 & 1.99 & 3.79 & 2.04 & 3.53 & 1.80 & 2.28 & 0.32 \\
        ~ & Same     & 3.61 & 1.80 & 4.48 & 2.03 & 4.07 & 1.80 & 3.67 & 0.16 \\
        ~ & Different& 3.16 & 2.24 & 3.48 & 2.06 & 3.17 & 1.78 & 0.90 & 0.64 \\

    \midrule
    \multicolumn{10}{l}{\textit{Intelligence}} \\
        ~ & Topic 1 & 5.32 & 1.74 & 5.55 & 1.35 & 5.60 & 0.89 & 0.08 & 0.96 \\
        ~ & Topic 2 & 5.23 & 1.48 & 5.55 & 1.44 & 5.57 & 1.07 & 1.41 & 0.49 \\
        ~ & Same     & 5.42 & 1.57 & 5.79 & 1.32 & 5.70 & 1.06 & 1.13 & 0.57 \\
        ~ & Different& 5.13 & 1.65 & 5.30 & 1.42 & 5.47 & 0.90 & 0.26 & 0.88 \\

    \midrule
    \multicolumn{10}{l}{\textit{Likeable}} \\
        ~ & Topic 1 & 5.29 & 1.83 & 5.33 & 1.45 & 5.30 & 1.32 & 0.42 & 0.81 \\
        ~ & Topic 2 & 5.32 & 1.70 & 5.39 & 1.52 & 5.70 & 0.99 & 0.23 & 0.89 \\
        ~ & Same     & 5.39 & 1.78 & 5.61 & 1.34 & 5.70 & 0.95 & 0.01 & 0.99 \\
        ~ & Different& 5.23 & 1.75 & 5.12 & 1.58 & 5.30 & 1.34 & 0.34 & 0.85 \\

    \bottomrule
    \end{tabularx}
\end{table*}
}

Table \ref{tab:general-impression} presents users’ general impressions of the agent(s), using the same analysis methods described in Section \ref{sec:statistic-analysis}.

\subsubsection{Qualitative}
We analyzed participants' impressions of the agent(s) from open-ended responses to assess overall usability and user satisfaction with the system: Participants generally described the agents' conversations as \textit{friendly, polite, and pleasant}. Although most participants recognized that the agents were AI-driven with scripted responses, they often used human-like descriptors, such as \textit{understanding, polite, respectful, and reasonable}. These impressions contributed to an engaging discussion atmosphere, which helped facilitate opinion changes. For instance, P75 from the 5-agent condition noted, \textit{"The agents were both informative and responsive, facilitating an engaging discussion. They presented ideas clearly and encouraged critical thinking."}

\subsection{Control Variables}
\label{app:control-variables}

Table \ref{tab:control-variable} reports the results for the control variables, which showed no significant differences across groups.

{\small
\begin{table*}
    \caption{Summary of Control Variables Across Experimental Conditions}
    \label{tab:control-variable}
    \centering
    \begin{tabularx}{\textwidth}{l l *{6}{>{\centering\arraybackslash}X} c c}
    \toprule
        \textbf{Item} & \textbf{Type} & \multicolumn{2}{c}{\textbf{1 Agent}} & \multicolumn{2}{c}{\textbf{3 Agents}} & \multicolumn{2}{c}{\textbf{5 Agents}} & H/F & p \\
        ~ & ~ & Mean & STD & Mean & STD & Mean & STD & ~ & ~ \\
    \midrule

    \multicolumn{10}{l}{\textit{Topic 1 Expertise}} \\
        ~ & Experience & 0.16 & 0.37 & 0.18 & 0.39 & 0.31 & 0.47 & 2.28 & 0.32 \\
        ~ & Familiarity & 2.65 & 0.66 & 2.91 & 0.77 & 2.79 & 0.62 & 3.66 & 0.16 \\
        ~ & Frequency   & 4.94 & 1.03 & 5.39 & 0.56 & 5.03 & 1.09 & 3.77 & 0.15 \\

    \midrule
    \multicolumn{10}{l}{\textit{Topic 2 Expertise}} \\
        ~ & Experience  & 0.74 & 0.44 & 0.94 & 0.24 & 0.79 & 0.41 & 4.68 & 0.10 \\
        ~ & Familiarity & 3.23 & 0.80 & 3.42 & 0.66 & 3.34 & 0.72 & 0.88 & 0.64 \\
        ~ & Frequency   & 3.81 & 1.51 & 4.52 & 1.15 & 4.55 & 1.18 & 4.88 & 0.09 \\

    \midrule
    \multicolumn{10}{l}{\textit{Conformity}} \\
        ~ & Positive & 5.87 & 0.73 & 5.37 & 1.06 & 5.25 & 1.11 & 4.80 & 0.10 \\
        ~ & Negative & 3.18 & 1.11 & 3.68 & 1.26 & 3.61 & 1.49 & 1.35 & 0.27 \\

    \midrule
    \multicolumn{10}{l}{\textit{Compliance}} \\
        ~ & Positive & 5.45 & 1.07 & 4.86 & 1.35 & 5.14 & 1.13 & 1.97 & 0.15 \\
        ~ & Negative & 4.40 & 1.18 & 4.37 & 1.23 & 4.56 & 0.94 & 0.25 & 0.78 \\

    \midrule
        AI Acceptance & - & 4.18 & 1.55 & 4.71 & 1.32 & 4.72 & 1.05 & 2.12 & 0.35 \\
        
    \bottomrule
    \end{tabularx}
\end{table*}
}

\subsection{Survey Items}
\label{app:survey}

\subsubsection{RQ1 - Opinion Change}
\begin{itemize}
    \item Topic 1 - \textit{Self-Driving Cars Should be allowed on Public Roads.} (Topic: \cite{govers2024ai})
    \begin{itemize}
        \item Self-driving cars should be allowed on public roads. 
        \item Allowing self-driving cars on public roads is a good idea. 
        \item Allowing self-driving cars on public roads has bad consequences. 
        \item Do you support or oppose allowing self-driving cars on public roads? (1=Strongly oppose, 6=Strongly support)
        \item If there was a referendum tomorrow on allowing self-driving cars on public roads, how likely is it that you would vote in favor? (1=Definitely would not, 6=Definitely would)
    \end{itemize}
    \item Topic 2 - \textit{Violent video games contribute to youth violence.} (Topic: \cite{yeo2024help})
    \begin{itemize}
        \item Violent video games contribute to youth violence. 
        \item Regulating violent video games to prevent youth violence is a good idea. 
        \item Allowing youth to play violent video games has bad consequences. 
        \item Do you support or oppose the regulation of violent video games to prevent youth violence? (1=Strongly oppose, 6=Strongly support)
        \item If there was a referendum tomorrow on regulating violent video games to prevent youth violence, how likely is it that you would vote in favor? (1=Definitely would not, 6=Definitely would)
    \end{itemize}

    \item Open-ended question - Explain why you chose your current stance.
\end{itemize}

\subsubsection{RQ2 - Social Influence}

\begin{itemize}
    \item Informational Influence
    \begin{itemize}
        \item My decision was influenced by the opinion of the agent(s).
        \item I was persuaded by the agent(s) and thus, I accepted the agent(s)' opinion.
    \end{itemize}
    \item Normative Influence
    \begin{itemize}
        \item I felt like I had to agree with the agent(s)' opinion during the discussion.
        \item I was not persuaded by agent(s)' opinion, but I accepted the agent(s)' opinion.
    \end{itemize}

     \item Open-ended question (General) - What do you think of the bots during the discussion? 
    \item Open-ended question (Accuracy) - Do you think the arguments presented by the agent(s) are accurate and convincing, and why?
    \item Open-ended question (Affiliation) - During the conversation, do you feel any pressure to agree with the agent(s)?

\end{itemize}

\subsubsection{Control Variables} 

\begin{itemize} 

  \item Topic 1 Expertise (Topic: \cite{govers2024ai})
    \begin{itemize}

        \item Experience - Have you ever been in a self-driving car? (0=No, 1=Yes)

        \item Frequency - How often do you drive? (1=Never, 6=Always)
            
        \item Familiarity - How familiar are you with self-driving cars? (1=Not familiar at all, 4=Very familiar)
        
    \end{itemize}

    \item Topic 2 Expertise (Topic: \cite{yeo2024help})
    \begin{itemize}
        \item Experience - Have you played violent video games before? (0=No, 1=Yes)
            
        \item Frequency - How often do you play video games? (1=Never, 6=Always)
            
        \item Familiarity - How familiar are you with violent video games? (1=Not familiar at all, 4=Very familiar)   
    \end{itemize}

    \item AI Acceptance \cite{pataranutaporn2023influencing}
    \begin{itemize} 
        \item There are many beneficial applications of AI.
        \item AI can help people feel happier.
        \item You want to use/interact with AI in daily life.
        \item AI can provide new economic opportunities.
        \item Society will benefit from AI.
        \item You love everything about AI.
        \item Some complex decisions should be left to AI.
        \item You would trust your life savings to an AI system.

    \end{itemize}

     \item Compliance Tendency \cite{gudjonsson1989compliance}
    \begin{itemize}
        \item Positive - I would never go along with what people tell me in order to please them.
        \item Positive -  I strongly resist being pressured to do things I don’t want to do.
        \item Positive - I am not too concerned about what people think of me.
        \item Negative - I would describe myself as a very obedient person.
        \item Negative - I generally tend to avoid confrontation with people.
        \item Negative - Disagreeing with people often takes more time than it is worth.
    \end{itemize}
    
    \item Conformity Tendency \cite{mehrabian1995basic}
    \begin{itemize}
        \item Positive - I don’t give in to others easily.
        \item Positive - I prefer to find my own way in life rather than find a group I can follow. 
        \item Positive - I am more independent than conforming in my ways.
        \item Negative - I often rely on, and act upon, the advice of others.
        \item Negative - Basically, my friends are the ones who decide what we do together. 
        \item Negative - If someone is very persuasive, I tend to change my opinion and go along with them.

    \end{itemize}
    
\end{itemize}

\subsubsection{Miscellaneous}

\begin{itemize}
    \item Impressions of agent(s) \cite{jakesch2023co, kim2024engaged}
    \begin{itemize}
        \item Understanding - The agent(s) understood what I wanted to say. 
        \item Expertise - The agent(s) were knowledgeable and had topic expertise.
        \item Balanced - The agent(s)' arguments were reasonable and balanced.
        \item Inspired - The agent(s) inspired or changed my thinking and argument.
        \item Intelligence - The agent(s) were intelligent. \cite{tanprasert2024debate}
        \item Likeble - The agent(s) were likeable.
    \end{itemize}

\end{itemize}

\begin{itemize}
    \item Multi-Choice Question (Attention Check) \cite{jakesch2023co}: What are we asking you to do in this task?
    
    \item Open-ended question (Study Purpose) - What do you think this study is trying to understand?

\end{itemize}

\subsection{Arguments}

\label{app:arguments}

Tables \ref{tab:topic1_arguments} and \ref{tab:topic2_arguments} list the arguments used in our system to guide the agents’ conversations.

{\small
\begin{table*}[t]
\centering
\caption{Support and Oppose Arguments for Topic 1 (Self-driving Cars)}
\label{tab:topic1_arguments}
\begin{tabularx}{\textwidth}{p{0.1\textwidth}X X}
\toprule
\textbf{Types} & \textbf{Support} & \textbf{Oppose} \\
\midrule
Topic 1 &
\textbf{More Time and Comfort} – Drivers can sit back and relax, take short breaks and devote their time to other things. If you're stuck in a long and tedious traffic jam, self-driving cars make things so much less stressful. &
\textbf{Technical Developments} – Technical developments are not yet perfect. Self-driving cars rely on complex algorithms and sensors that can sometimes fail, leading to accidents. \\
& \textbf{Safety} – Self-driving cars need advanced sensors and algorithms to function correctly. Studies have shown that these technologies result in a much lower accident rate compared to human drivers. &
\textbf{Vehicle Communication} – Self-driving cars need to communicate with each other and with traffic infrastructure to function correctly. Any disruption in this communication could cause severe problems. \\
& \textbf{Efficiency in Traffic} – Self-driving cars also promise more efficiency in traffic. They can communicate with each other to optimize traffic flow, reducing congestion and travel time. &
\textbf{Surveillance} – Self-driving cars raise significant surveillance issues. They collect extensive data about their passengers and surroundings, which could be misused or hacked. \\
& \textbf{Accessibility} – For people who are unable to drive due to age, disability, or other reasons, self-driving cars could provide newfound independence and mobility. &
\textbf{Legal} – Imagine there is an accident involving a self-driving car and a human. Who is responsible for such a situation? It's hard to ask either the company or the driver to take responsibility. \\
& \textbf{Navigation} – Self-driving cars excel in navigating complex routes. Their advanced systems can interpret GPS instructions with high accuracy. It'll even help you find a place to park when you get there. &
\textbf{Mixed Traffic} – The coexistence of human-driven and self-driving cars could create complex situations on the road, leading to potential accidents. \\
\bottomrule
\end{tabularx}
\end{table*}
}

{
\small
\begin{table*}[t]
\centering
\caption{Support and Oppose Arguments for Topic 2 (Violent Video Games)}
\label{tab:topic2_arguments}
\begin{tabularx}{\textwidth}{p{0.1\textwidth}X X}
\toprule
\textbf{Types} & \textbf{Support} & \textbf{Oppose} \\
\midrule
Topic 2 &
\textbf{Aggression} – Playing violent video games can cause more aggression, bullying, and fighting among youth. &
\textbf{Exploring Consequences} – Experiencing violence in games can actually have a positive effect on children, as it lets them explore consequences of violent actions in a safe space. \\
& \textbf{Desensitization} – Simulating violence, such as shooting guns and hand-to-hand combat in video games, can desensitize youth to real-life violence. &
\textbf{Moral Development} – Games let youth develop their own sense of right or wrong, and can help them release stress harmlessly. \\
& \textbf{Mass Shooters} – Many perpetrators of mass shootings have been found to have played violent video games. &
\textbf{Positive Effects} – Studies show violent video games can have positive effects on kindness, civic engagement, and prosocial behaviors. \\
& \textbf{Psychological Impacts} – Violent video games can cause reduced empathy and increased likelihood of aggression, especially with other risk factors. &
\textbf{Scapegoat for Societal Issues} – These games are often blamed for violence instead of deeper issues like lack of support systems or poor education. \\
& \textbf{Reality-Fantasy Confusion} – Children may imitate violent characters and struggle to distinguish fantasy from reality. &
\textbf{Weak Evidence} – There is little evidence linking violent games to real-world violence; other societal issues are more important to address. \\
\bottomrule
\end{tabularx}
\end{table*}
}

\subsection{Prompts}

\subsubsection{LLM Prompt to Enhance Discussion}
\label{app:prompts}

\begin{quote}
\textit{You are having a conversation with the user on "whether self-driving cars should be allowed on all roads". Your stance is supporting the topic.  Specifically, you asked the user: "Would you enjoy having more time to yourself if you didn't have to focus on driving?". Based on user's input, first give a reply of around 20 words acknowledging the user's opinion on what they think, then ask the user to share more opinions on the topic.}
\end{quote}

\subsubsection{LLM Prompt to Acknowledge User Opinion}

Sample prompt to generate a message that acknowledges the user's opinion on a topic:
\begin{quote}
\textit{You are talking to a user on "whether self-driving cars should be allowed on all roads". Your stance is \textbf{supporting} the topic. You just shared your opinion on how self-driving cars promote accessibility to disability and asked if the user agrees with these reasons for having self-driving cars. Give a reply of around 20 words acknowledging the user's opinion on what they like and/or don't like.}
\end{quote}

After this, in the next message they sent, the agent(s) continued to follow the script and express their own stance (either for or against).

\end{document}